\documentclass[lettersize,journal]{IEEEtran}
\usepackage{amsmath,amsfonts}
\usepackage{array}

\usepackage{textcomp}
\usepackage{stfloats}
\usepackage{verbatim}
\usepackage{cite}
\usepackage{bm}
\usepackage{times}
\usepackage{epsfig}
\usepackage{bbding}
\usepackage{subfigure}
\usepackage{graphicx}
\usepackage{amsthm}
\usepackage{calc}
\usepackage[ruled,linesnumbered]{algorithm2e}
\usepackage{bbm}
\usepackage{color}
\usepackage{xcolor}
\usepackage{url}
\usepackage{enumitem}
\usepackage{float}
\usepackage{multirow}
\usepackage{multicol}
\usepackage{mathrsfs}

\newtheorem{Lemma}{Lemma}
\newtheorem{Theorem}{Theorem}

\newcommand{\ie}{\textit{i}.\textit{e}.}

\graphicspath{{./images/}}
\begin{document}

\title{Quaternion Infrared Visible Image Fusion}

\author{Weihua Yang and Yicong Zhou,~\IEEEmembership{Senior Member,~IEEE}
\thanks{This work was funded by the Science and Technology Development Fund, Macau SAR (File no. 0049/2022/A1, 0050/2024/AGJ), by the University of Macau (File no. MYRG-GRG2024-00181-FST). (Corresponding author: Yicong Zhou.)}
\thanks{Weihua Yang and Yicong Zhou are with the Department of Computer and
Information Science, University of Macau, Macau 999078, China. (e-mail:
weihuayang.um@gmail.com; yicongzhou@um.edu.mo)}}

\markboth{IEEE TRANSACTIONS ON IMAGE PROCESSING}%
{Shell \MakeLowercase{\textit{et al.}}: Quaternion Multi-focus Color Image Fusion}
\IEEEpubid{0000--0000/00\$00.00~\copyright~2025 IEEE}

\maketitle

\begin{abstract}
Visible images provide rich details and color information only under well-lighted conditions while infrared images effectively highlight thermal targets under challenging conditions such as low visibility and adverse weather. Infrared-visible image fusion aims to integrate complementary information from infrared and visible images to generate a high-quality fused image. Existing methods exhibit critical limitations such as neglecting color structure information in visible images and performance degradation when processing low-quality color-visible inputs.
To address these issues, we propose a quaternion infrared-visible image fusion (QIVIF) framework to generate high-quality fused images completely in the quaternion domain. QIVIF proposes a quaternion low-visibility feature learning model to adaptively extract salient thermal targets and fine-grained texture details from input infrared and visible images respectively under diverse degraded conditions. QIVIF then develops a quaternion adaptive unsharp masking method to adaptively improve high-frequency feature enhancement with balanced illumination. QIVIF further proposes a quaternion hierarchical Bayesian fusion model to integrate infrared saliency and enhanced visible details to obtain high-quality fused images.
Extensive experiments across diverse datasets demonstrate that our QIVIF surpasses state-of-the-art methods under challenging low-visibility conditions.
\end{abstract}


\section{Introduction}
\begin{figure}[hbtp]
\centering
\includegraphics[width=1\columnwidth]{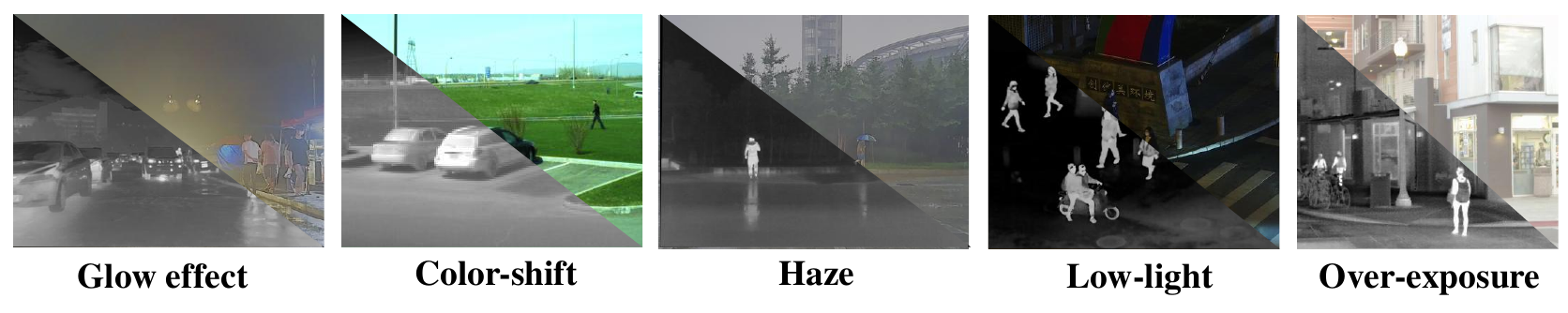}
\caption{Samples pairs of the infrared and the visible from the challenging low-visibility scenarios including glow effects, color shifts, haze, low light and over-exposure conditions. }
\label{fig:data}
\end{figure}

Infrared-visible image fusion (IVIF) techniques enable complementary features of infrared and visible images to be combined into a fused image. It significantly facilitates subsequent applications such as object detection \cite{wang2023interactively,sun2021fusion}, target tracking \cite{yuan2023thermal,zhang2019siamft}  and biometric recognition \cite{li2024hand,ariffin2017can}.
As seen in Fig. \ref{fig:data}, visible images exhibit reflection information while infrared imaging captures the thermal radiation of targets and low-texture background without the negative impacts of atmospheric light.

\begin{figure}[hbtp]
\centering
\includegraphics[width=1\columnwidth]{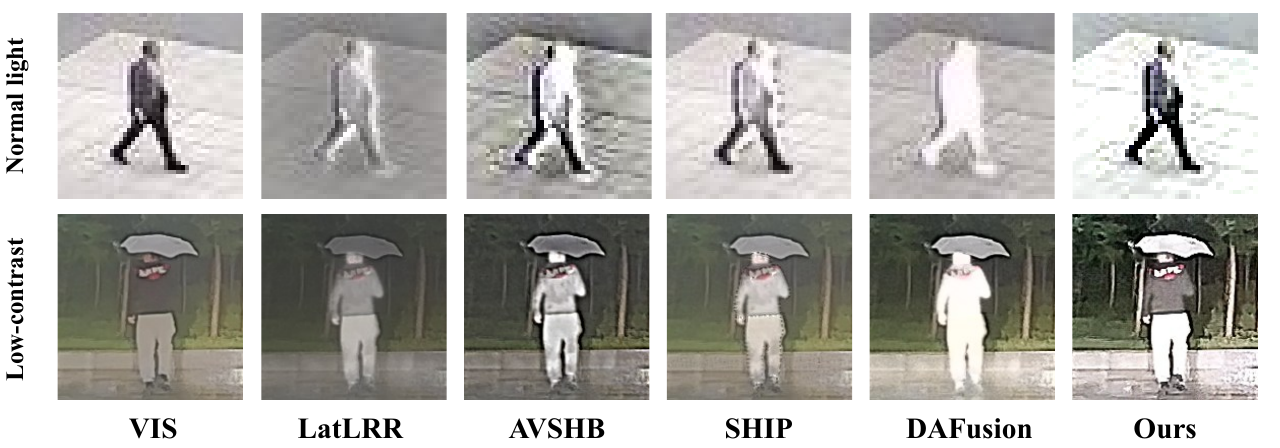}
\caption{ Visual Comparison of various IVIF algorithms in normal and low visibility conditions. VIS denotes the visible images. LatLRR \cite{wang2020latent} and AVSHB \cite{fu2022adaptive} correspond to traditional model-based IVIF methods. SHIP \cite{zheng2024probing} and DAFusion \cite{wang2025degradation} are deep-learning-based methods. Compared to deep-learning-based methods, LatLRR and AVSHB inject more grayscale pixels of the infrared and exhibit a low-contrast background particular in the low-visibility condition.  }
\label{fig:comparison-modeltype}
\end{figure}

Existing IVIF methods can be broadly categorized into traditional model-based and deep-learning-based ones. 
Recent advances in deep-learning-based methods leverage diverse architectures including generative adversarial networks \cite{liu2022target,yang2021infrared}, and vision transformers \cite{ma2022swinfusion}. They have achieved competitive performance. However, these deep-learning-based IVIF methods focus on preserving illumination-based thermal saliency in infrared images while ignoring color structure and details in the visible images \cite{yue2023dif}. For instance, when processing visible images, many deep-learning-based IVIF methods \cite{tang2023divfusion,zheng2024probing} convert RGB images into YCbCr space and retain only the luminance (Y) channel for image fusion while discarding chrominance (Cb/Cr) channels during feature extraction and fusion. This transformation inherently neglects cross-channel interactions and often results in fused images with color casts or desaturation \cite{yue2023dif}. To circumvent degraded scenarios, recent deep-learning-based IVIF methods adopt auxiliary task-guided learning by integrating illumination-based enhancement modules \cite{tang2023divfusion,wang2025degradation} to enhance feature extraction and fused images with task-specific priors. 
However, their fused images often exhibit obscured critical details that hinder human perception. As seen in Fig. \ref{fig:comparison-modeltype}, despite high saliency in fused images, deep-learning-based methods fail to effectively maintain fine-grained details of visible images.

\IEEEpubidadjcol On the other hand, traditional model-based IVIF methods are inherently training-free, bypassing the need for data-driven feature representation learning. However, traditional model-based IVIF methods are predominantly designed for fusing single-channel infrared and grayscale visible image pairs. When applied to color visible images, traditional model-based methods often rely on channel-wise fusion strategies that treat each color channel as a grayscale image and fuse them with the infrared one-by-one. For example, \cite{fu2022adaptive} designed a two-scale decomposition model to obtain the base (low-frequency) and detail (high-frequency) layers from grayscale pairs. It independently fuses these layers across infrared and visible images via heuristic rules and generates fused images channel-by-channel \cite{tang2024egefusion}. The fused color image is synthesized by concatenating three independently fused channels. Such channel-wise processing fails to consider cross-channel interactions of color images and leads to color distortion \cite{yu2013quaternion}.
Traditional IVIF methods leverage multi-scale feature extraction and fusion (e.g., edge-preserving filters, gradient operators, low-rank representation) to enhance edges and gradients common to both infrared and visible images \cite{wang2020latent,li2020mdlatlrr,fu2022adaptive}.
However, these methods insufficiently adapt to the modality-specific characteristics of the infrared and visible images and may result in insufficient or redundant feature enhancement as seen in Fig. \ref{fig:comparison-modeltype}. Traditional methods struggle to disentangle scenario-specific details from degraded visible images and often overcompensate by fusing redundant thermal information into the fused images \cite{tang2023divfusion}. This imbalance leads to low-visibility where fused images exhibit obvious shadows, washed-out textures or loss of chromatic fidelity for human perception (see Fig. \ref{fig:comparison-modeltype}).

Quaternion representation is an effective tool for color image processing, encoding a color image as a pure quaternion matrix. It preserves the inter-relationships between the red, green, and blue channels of color images \cite{miao2020quaternion,jia2025new} and fully utilizes the highly correlated color information. This provides better performance compared to real-valued methods \cite{huang2021quaternion}. Specifically, quaternion-based low-rank regularization leverages the algebraic properties of quaternions while enforcing low-rank constraints. This approach has been widely applied in various color image processing tasks, such as color image recovery \cite{miao2023quaternion}, color image denoising \cite{chen2019low}, and color image stitching \cite{li2022automatic}. Furthermore, \cite{lin2024loquat} demonstrated that quaternion representation enhances the exploration of low-rank properties and provides superior quality performance.

Taking advantage of the quaternion representation, this paper introduces a unified quaternion infrared visible fusion framework to simultaneously address the above fusion challenges.
Our main contributions are presented as follows:
\begin{itemize}
    \item We propose a quaternion image fusion (QIVIF) framework to address multiple low-contrast scenarios (e.g., glow effect, color-shift, haze, low-light and over-exposure). To the best knowledge, our QIVIF is the first to perform infrared and visible image fusion \textbf{completely in the quaternion domain}.
    \item QIVIF proposes a unified quaternion low-visibility feature learning model to adaptively disentangle infrared saliency (thermal targets) and visible details (textures, edges) while preserving highly correlated color information under diverse low-visibility scenarios.
    \item QIVIF develops a quaternion adaptive unsharp masking method to obtain a promising detail-enhanced visible image with balanced illumination. We are the first to bring the unsharp masking concept to the quaternion domain for color image enhancement. 
    \item QIVIF proposes a quaternion hierarchical Bayesian fusion model to integrate infrared saliency and enhanced visible details using adaptive feature weighting. This ensures trade-offs between high saliency of thermal targets and high visibility for human perception in fused images.
    \item Extensive experiments on various datasets demonstrate that our QIVIF outperforms the state-of-the-art methods.
\end{itemize}
The rest of this paper is organized as follows: Section \ref{overview} presents the preliminaries of quaternion representation. Section \ref{framework:QIVIF} introduces our framework in detail. Section \ref{experiments} presents the experiments and comparisons. Finally, Section \ref{conclusion} gives the conclusions.
\section{Preliminaries} \label{overview}
Quaternions provide a holistic color image processing by encapsulating color channels within a single algebraic structure. Below is a detailed introduction to quaternion algebra and its application in color image representation. 
\subsection{Quaternion algebra and representation}
A quaternion scalar denoted as $\dot{q}$ $\in \mathbbm{H}$  is a hypercomplex number with one real part and three imaginary parts. It is defined as
$\dot{q}=a+b \boldsymbol{i}+c \boldsymbol{j}+ d  \boldsymbol{k},$ where $a,b,c,d \in \mathbbm{R}$ and $\boldsymbol{i},\boldsymbol{j},\boldsymbol{k}$ are imaginary units satisfying:
$\boldsymbol{i}^2=\boldsymbol{j}^2=\boldsymbol{k}^2=\boldsymbol{i}\boldsymbol{j}\boldsymbol{k}=-1$.

Quaternion addition and subtraction follow a component-wise rule as that in the complex space. However, quaternion multiplication is generally non-commutative and the multiplication rule is defined as
$\boldsymbol{i}\boldsymbol{j}=-\boldsymbol{j}\boldsymbol{i}=\boldsymbol{k},\boldsymbol{j}\boldsymbol{k}=-\boldsymbol{k}\boldsymbol{j}=\boldsymbol{i},\boldsymbol{k}\boldsymbol{i}=-\boldsymbol{i}\boldsymbol{k}=\boldsymbol{j}.
$
Quaternion conjugate and modulus are key operations defined below,
\begin{equation*}
    \overline{\dot{q}}=a-b \boldsymbol{i}-c \boldsymbol{j}- d  \boldsymbol{k}, \vert\dot{q} \vert=\sqrt{a^2+b^2+c^2+d^2}.
\end{equation*}
\subsection{Quaternion representation}
\begin{figure}[hbtp]
\centering
\includegraphics[width=1\columnwidth]{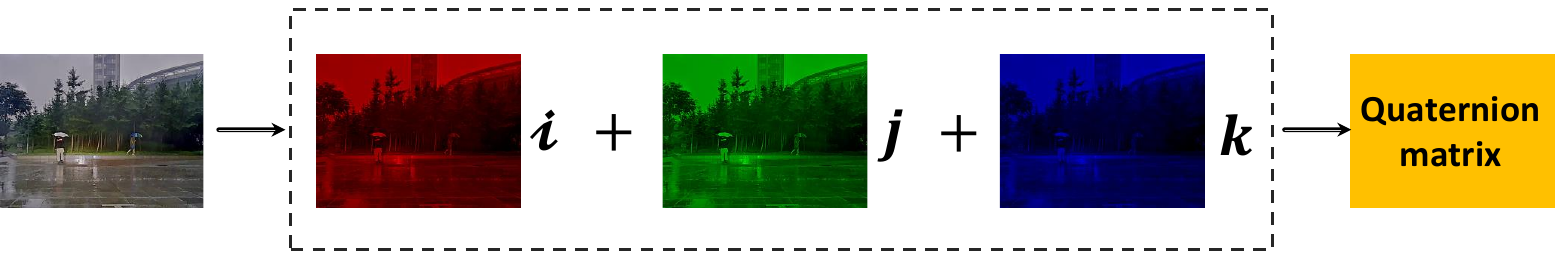}
\caption{Quaternion representation of a color image. We encode the three channels of the color image into a pure quaternion matrix.}
\label{fig:QR}
\end{figure}

Quaternions provide mathematically elegant color image processing by unifying RGB channels into a single hypercomplex entity. For color image representation, each color pixel q with its RGB values $(r,g,b)$ is encoded as a pure quaternion scalar
$
    \dot{q}=r \boldsymbol{i}+g \boldsymbol{j}+b  \boldsymbol{k}.
$
As shown in Fig. \ref{fig:QR}, three color channels of a color image $\boldsymbol{\mathrm{A}}$ are represented as two-dimensional pure quaternion matrix $\dot{\boldsymbol{\mathrm{A}}}$.

For a quaternion matrix $\dot{\boldsymbol{\mathrm{A}}}$ $\in {\mathbbm{H}}^{H\times W}$ with each element denoted as $\dot{a}_{i,j}$, its conjugate transpose is denoted as $\dot{{\boldsymbol{\mathrm{A}}}}^{\mathrm{H}}=(\overline{\dot{a}_{j,i}}) \in \mathbbm{H}^{W\times H}$. The ${\ell}_1$ and $Frobenius$ norms are computed as $\small {\Vert \dot{\boldsymbol{\mathrm{A}}} \Vert}_1=\sum_{i=1}^{H}\sum_{j=1}^{W}{\vert \dot{a}_{i,j}\vert}, {\Vert \dot{\boldsymbol{\mathrm{A}}} \Vert}_F={(\sum_{i=1}^{H}\sum_{j=1}^{W}{{\vert \dot{a}_{i,j}\vert}^2})}^{\frac{1}{2}}$.

Spatial derivatives in $x$ and $y$ directions can be computed in the quaternion domain. The gradient magnitude combines information from all channels to produce consistent color edges. The first-order quaternion gradients \cite{huang2023review} can be defined as $\frac{\partial \dot{\boldsymbol{\mathrm{I}}}}{\partial x}$ and $\frac{\partial \dot{\boldsymbol{\mathrm{I}}}}{\partial y}$ at a vertical or horizontal direction. The second-order gradients can be defined as $\frac{{\partial}^2 \dot{\boldsymbol{\mathrm{I}}}}{{\partial}^2 x}$, $\frac{{\partial}^2 \dot{\boldsymbol{\mathrm{I}}}}{{\partial}^2 y}$ and $\frac{{\partial}^2 \dot{\boldsymbol{\mathrm{I}}}}{{\partial} x {\partial} y}$.

\subsection{Quaternion low rank regularization}
Let a quaternion matrix $\dot{\boldsymbol{\mathrm{X}}}$ $\in {\mathbbm{H}}^{H\times W}$ be constrained to a low-rank structure. To approximate its solution, the quaternion nuclear norm as a convex surrogate for rank minimization is defined as ${\|\dot{\boldsymbol{\mathrm{X}}}\|}_{*}=\sum_i^r {\sigma}_i(\dot{\boldsymbol{\mathrm{X}}})$ where ${\sigma}_i(\dot{\boldsymbol{\mathrm{X}}})$ denote the real non-negative singular values of $\dot{\boldsymbol{\mathrm{X}}}$ obtained from quaternion singular value decomposition (QSVD) \cite{xiao2018two}.QSVD factorizes ${\sigma}_i(\dot{\boldsymbol{\mathrm{X}}})$ into quaternion-valued unitary matrices and a diagonal matrix of real singular values.

 However, the quaternion nuclear norm often yields suboptimal rank approximations due to its uniform penalization of all singular values \cite{huang2022quaternion}. To address this, weighted Schatten-$p$ norm and partial sum of singular values are extended to the quaternion domain.

\begin{Lemma}
(Quaternion weighted Schatten-$p$ norm \cite{miao2020quaternion}) For any $\lambda \geq 0$, quaternion matrix $\dot{\boldsymbol{\mathrm{Y}}}$ and $\dot{\boldsymbol{\mathrm{X}}}$ $\in \mathbbm{H}^{H\times W}$ with the rank of $r$, then the quaternion weighted Schatten p-norm problem ($0<p<1$) can be defined as:
\begin{align}
    \mathop{\arg\min}_{\dot{\boldsymbol{\mathrm{X}}}}\ \  \frac{1}{2}{\Vert \dot{\boldsymbol{\mathrm{Y}}}-\dot{\boldsymbol{\mathrm{X}}} \Vert}_{F}^2+ \lambda{\Vert \dot{\boldsymbol{\mathrm{X}}} \Vert}_{w,S_p}^p, \label{eq2.2.1}
\end{align}
with the given $p$ and $w$, there exists a specific threshold ($p<1$) ${\tau}_p^{GST}(\lambda w)={(2\lambda w(1-p))}^{\frac{1}{2-p}}+\lambda w p{(2\lambda w(1-p))}^{\frac{p-1}{2-p}}$. Then we have two conclusions. 
(1) When ${\sigma}_i(\dot{\boldsymbol{\mathrm{Y}}}) \leq {\tau}_p^{GST}(\lambda w)$, the optimal solution of equation \eqref{eq2.2.1} is ${\sigma}_i(\dot{\boldsymbol{\mathrm{X}}})=0$.
(2) When ${\sigma}_i(\dot{\boldsymbol{\mathrm{Y}}}) > {\tau}_p^{GST}(\lambda w)$, the optimal solution of equation \eqref{eq2.2.1} is ${\sigma}_i(\dot{\boldsymbol{\mathrm{X}}})-{\sigma}_i(\dot{\boldsymbol{\mathrm{Y}}})+\lambda wp{({\sigma}_i(\dot{\boldsymbol{\mathrm{X}}}))}^{p-1}$.
\label{lma2.2.2}
\end{Lemma}

\begin{Lemma}
(Quaternion partial sum of singular values \cite{oh2015partial})
For any $\lambda \geq 0$, quaternion matrix $\dot{\boldsymbol{\mathrm{Y}}}$ and $\dot{\boldsymbol{\mathrm{X}}}$ $\in \mathbbm{H}^{H\times W}$ with the rank of $r$, then the minimization problem of the partial sum of quaternion singular values can be defined as:
\begin{align}
    \mathop{\arg\min}_{\dot{\boldsymbol{\mathrm{X}}}}\ \  \frac{1}{2}{\Vert \dot{\boldsymbol{\mathrm{Y}}}-\dot{\boldsymbol{\mathrm{X}}} \Vert}_{F}^2+ \lambda{\Vert \dot{\boldsymbol{\mathrm{X}}} \Vert}_{p=n} \label{eq2.2.2}
\end{align}
the closed form solution of this equation is $\dot{\boldsymbol{\mathrm{U}}}_1 {\Sigma}_1\dot{\boldsymbol{\mathrm{V}}}_1^{\mathrm{H}}+\dot{\boldsymbol{\mathrm{U}}}_2 S_\lambda({\Sigma}_2)\dot{\boldsymbol{\mathrm{V}}}_2^\mathrm{H}$ using QSVD of $\dot{\boldsymbol{\mathrm{Y}}}$, where ${\Sigma}_1 $ is a diagonal real matrix with singular values $diag({\sigma}_1, {\sigma}_2,\cdots, {\sigma}_n )$, ${\Sigma}_2 $ is a diagonal real matrix with singular values $diag({\sigma}_{n+1}, {\sigma}_{n+2},\cdots, {\sigma}_r )$ and $S_\lambda(\cdot)$ denotes the soft thresholding operator with parameter $\lambda$.
\label{lma2.2.3}
\end{Lemma}
\begin{figure}[hbtp]
	\centering
	\includegraphics[width=1\columnwidth]{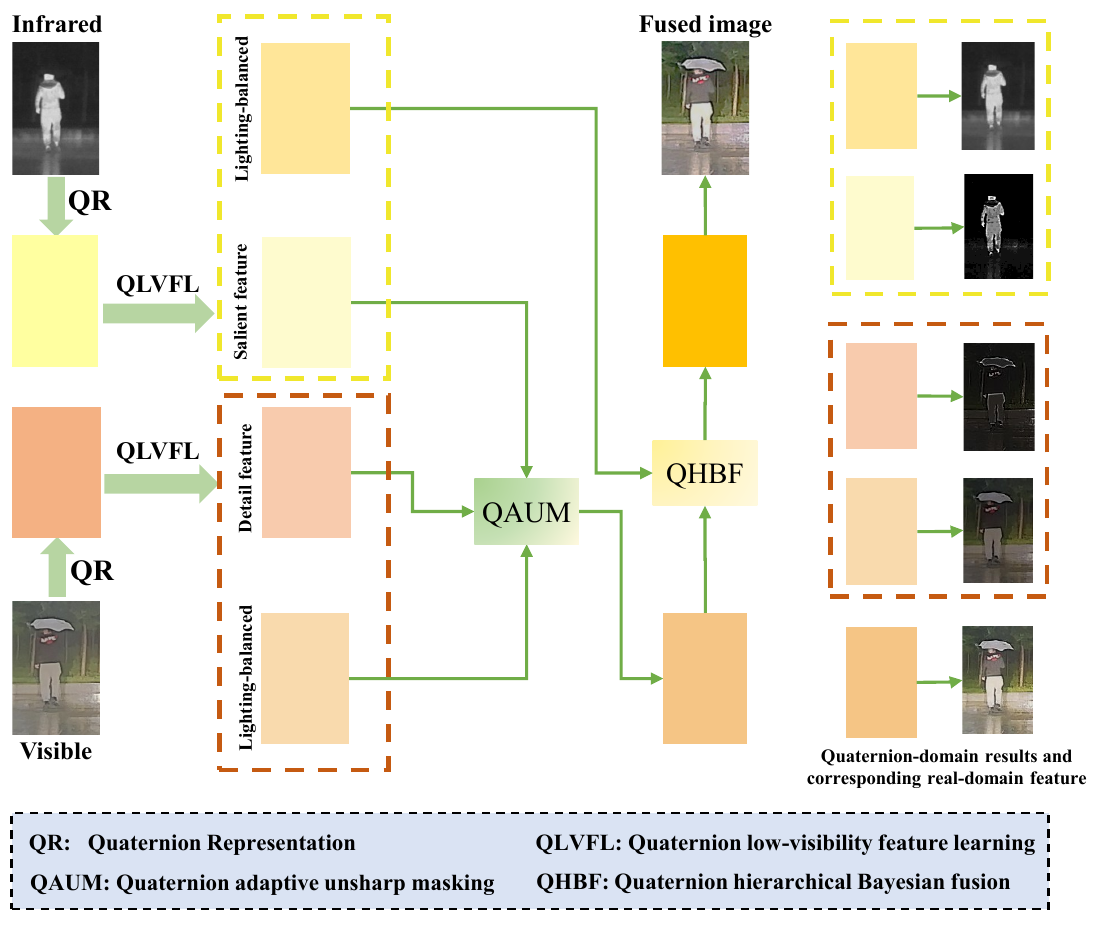}\\
	\caption{Flowchart of our quaternion infrared visible image fusion. }\label{framework-flowchart}
\end{figure}
\section{Proposed Framework} \label{framework:QIVIF}
\subsection{Overview}
 Our quaternion infrared visible image fusion (QIVIF) framework proposes a pixel-level feature extraction and fusion pipeline as illustrated in Fig. \ref{framework-flowchart}. To holistically model the cross-channel relationships within a color visible image, QIVIF first encodes visible and infrared images as pure quaternion matrices using quaternion representation. The three RGB channels of the visible image are mapped to the three imaginary components of a quaternion matrix to preserve high-correlated color information. The single-channel infrared image is replicated across three dimensions to construct a compatible quaternion structure. This enables unified feature representation of both modalities within a shared framework. To address feature extraction under diverse low-visibility scenarios, QIVIF introduces a unified quaternion low-visibility feature learning model. This model derives lighting-suppressed quaternion representations and disentangles modality-specific features for the infrared and visible quaternion representations respectively. Furthermore, QIVIF proposes a quaternion adaptive unsharp masking method to enhance fine-grained details and target saliency and reconstruct a sharpened visible quaternion representation with balanced illumination inspired by \cite{panetta2011nonlinear}. Finally, QIVIF introduces a quaternion hierarchical Bayesian fusion model to merge the sharpened visible and lighting-suppressed infrared quaternion representations. This model effectively generates fused images with critical image details and high-contrast visibility.

\subsection{Quaternion low-visibility feature learning}
\label{QLRD}
To handle degraded color-visible images and learn modality-specific features for both infrared and visible images, this subsection proposes a unified quaternion low-visibility feature learning (QLVFL) model. It consists of a quaternion lighting suppression module to remove various degradation effects and a quaternion low-rank decomposition module to learn the thermal saliency and fine-grained image details for infrared and visible images respectively.
\subsubsection{Quaternion lighting suppression}
Visible images primarily capture reflected light and offer an intuitive representation of objects with abundant details. However, they are limited in overcoming environmental disturbances \cite{fu2022adaptive}.
The quaternion lighting suppression (QLS) module aims to eliminate the negative impacts of diverse degraded effects including glow effects, color shifts, haze, low-light and overexposure conditions. This module assumes that the low-quality image can be decomposed into a low-illumination layer and an over-smooth and bright layer representing the negative impacts of the scene.
\begin{equation*}
    \dot{\boldsymbol{\mathrm{L}}}(x)=\dot{\boldsymbol{\mathrm{I}}}(x)+\dot{\boldsymbol{\mathrm{G}}}(x);
\end{equation*}
Considering the brightness decreases
gradually and smoothly, we exploit this smoothness
attribute and employ the method of target layer separation for scenes where one layer is significantly smoother than the other \cite{li2015nighttime}. The core is to remove the bright regions with low gradients and retain large gradients for the separated layer $\dot{\boldsymbol{\mathrm{I}}}$. The formulation is presented as follows:
\begin{equation}
    \begin{aligned}
    &\mathop{\arg\min}_{\dot{\boldsymbol{\mathrm{I}}}}\ \ {\|H(\nabla \dot{\boldsymbol{\mathrm{I}}}) \|}_1 +\lambda{\|\triangle \dot{\boldsymbol{\mathrm{I}}}-\triangle\dot{\boldsymbol{\mathrm{L}}}\|}_F^2,\\
    &s.t.\ \ 0\leq b(\dot{\boldsymbol{\mathrm{I}}})\leq b(\dot{\boldsymbol{\mathrm{L}}}), \dot{\boldsymbol{\mathrm{G}}}=H(\nabla \dot{\boldsymbol{\mathrm{I}}})
    \end{aligned} \label{eq2.2.14}
\end{equation}
where $\nabla$ denotes the first-order gradient filter, and $\triangle$ denotes the second-order gradient filter. $b(\dot{\boldsymbol{\mathrm{I}}})$ is a function to extract the intensity component of quaternion matrix $\dot{\boldsymbol{\mathrm{I}}}$ (see details in supplementary material). The function $H(\dot{\boldsymbol{\mathrm{I}}})$ is to minimize the small values of $\dot{\boldsymbol{\mathrm{I}}}$ as following equation:
\begin{equation}
    H(\dot{\boldsymbol{\mathrm{X}}})(s,t)=\left \{ 
    \begin{aligned}
        &\dot{\boldsymbol{\mathrm{X}}}(s,t),& &\| \dot{\boldsymbol{\mathrm{X}}}(s,t)\|>\tau\\
        &0,& &{\rm otherwise}.
    \end{aligned}\label{eq2.2.15}
    \right.
\end{equation}

\subsubsection{Quaternion low-rank decomposition}

Given an observed three-dimensional matrix $\boldsymbol{\mathrm{I}}$ $\in \mathbbm{R}^{H \times W \times 3}$ (\ie, a visible image) and its quaternion representation can be denoted as $\dot{\boldsymbol{\mathrm{I}}}$. For the infrared image $\boldsymbol{\mathrm{I}}_f$ $\in \mathbbm{R}^{H \times W}$, we first replicate the grayscale channel to construct a three-dimensional matrix then we apply the quaternion representation in the new infrared image with the same three channels.
Considering the smooth property in the background area and sparsity of the detail layer, we propose a quaternion low-rank decomposition module (QLRD). It is assumed to be decomposed into a structure layer $\dot{\boldsymbol{\mathrm{Z}}}$ and a detail layer $\dot{\boldsymbol{\mathrm{D}}}$ with weighted low rank prior as.
\begin{figure}
	\centering
	\includegraphics[width=1\columnwidth]{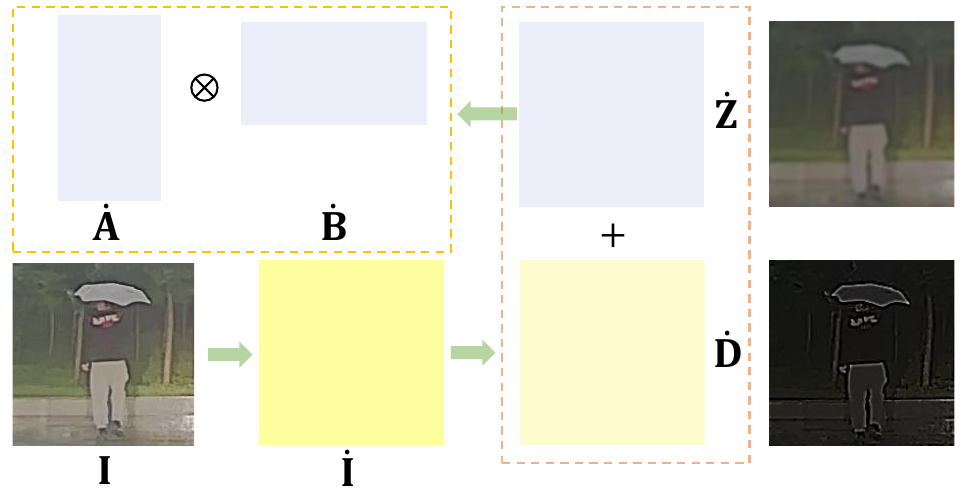}\\
	\caption{Decomposition results of the infrared and the visible using QLRD model. The QLRD model decomposes the given input quaternion representation into a smooth structure feature layer and a salient target feature layer with rich details.  }\label{decomposition-flowchart}
\end{figure}
\begin{equation}
    \begin{aligned}
    &\mathop{\arg\min}_{\dot{\boldsymbol{\mathrm{Z}}},\dot{\boldsymbol{\mathrm{D}}}} \ \ {\|\dot{\boldsymbol{\mathrm{Z}}} \|}_{w,S_p,p_1=n}^p+\beta{\| \dot{\boldsymbol{\mathrm{D}}}\|}_1 +\gamma{\|\dot{\boldsymbol{\mathrm{E}}}\|}_F^2 \\
    &s.t. \ \ \dot{\boldsymbol{\mathrm{I}}}=\dot{\boldsymbol{\mathrm{Z}}}+\dot{\boldsymbol{\mathrm{D}}}+\dot{\boldsymbol{\mathrm{E}}}, \label{eq2.2.5}
    \end{aligned}
\end{equation}
where $\dot{\boldsymbol{\mathrm{Z}}}$ $\in \mathbbm{H}^{H\times W}$ is constrained in quaternion weighted Schatten $p$-norm and weighted diagonal matrix $\boldsymbol{\mathrm{W}}$. To better relax $\dot{\boldsymbol{\mathrm{Z}}}$ in this nonconvex term, the partial sum of singular values is utilized to minimize residual rank when $p_1>n$. Since the non-convex surrogate for the rank function, the
Schatten-$p$ norm with $0 < p < 1$ makes a closer approximation to the rank function than the nuclear norm.
\begin{Lemma}
(Quaternion singular value inequality of bilinear factorization \cite{miao2020quaternion}) Let two quaternion matrices $\dot{\boldsymbol{\mathrm{X}}}$ $\in \mathbbm{H}^{H\times r} $ and $\dot{\boldsymbol{\mathrm{Y}}}$ $\in \mathbbm{H}^{r\times W} $ be given and let $p>0$. $K$ is denoted as $min(H,W,r)$, The inequality holds for the decreasingly ordered singular values of $\dot{\boldsymbol{\mathrm{X}}}$,  $\dot{\boldsymbol{\mathrm{Y}}}$ and $\dot{\boldsymbol{\mathrm{X}}}\dot{\boldsymbol{\mathrm{Y}}}$.
\begin{align}
    \sum_{k}^K {{\sigma}_k^p(\dot{\boldsymbol{\mathrm{X}}}\dot{\boldsymbol{\mathrm{Y}}})} \leq \sum_{k}^K {{\sigma}_k^p(\dot{\boldsymbol{\mathrm{X}}}) {\sigma}_k^p(\dot{\boldsymbol{\mathrm{Y}}})} \label{eq2.2.3}
\end{align}
where ${\sigma}_k(\dot{\boldsymbol{\mathrm{X}}})$ is the $k$-$th$ singular value of $\dot{\boldsymbol{\mathrm{X}}}$.
\label{lma2.2.4}
\end{Lemma}

Inspired by \cite{miao2020quaternion}, for $\dot{\boldsymbol{\mathrm{Z}}}$ $\in \mathbbm{H}^{H\times W}$, it is constrained with rank $r< min(H,W)$ and can be decomposed into two smaller quaternion factor matrices $\dot{\boldsymbol{\mathrm{A}}}$ $\in \mathbbm{H}^{H\times r}$ and $\dot{\boldsymbol{\mathrm{B}}}$ $\in \mathbbm{H}^{r\times W}$  such that $\dot{\boldsymbol{\mathrm{Z}}}=\dot{\boldsymbol{\mathrm{A}}}\dot{\boldsymbol{\mathrm{B}}}$.
Lemma \ref{lma2.2.4} allows us to make a factorization of
larger quaternion matrices for arbitrary $p$ quasi-norm. 

Based on this, we develop a new inequality of bilinear factorization under quaternion weighted Schatten $p$-norm in the following Theorem \ref{the2.1}.

\begin{Theorem}
Let a quaternion matrix $\dot{\boldsymbol{\mathrm{X}}}$ $\in \mathbbm{H}^{h\times w} $ and a diagonal weight matrix $\boldsymbol{\mathrm{W}}$ be given, then the decomposed quaternion matrices $\dot{\boldsymbol{\mathrm{A}}}$ $\in \mathbbm{H}^{h\times r} $ and $\dot{\boldsymbol{\mathrm{B}}}$ $\in \mathbbm{H}^{r\times W} $ can be obtained. Let $p>0$. $K$ is denoted as $min(H,W,r)$. The following inequality holds for the decreasingly ordered singular values of $\dot{\boldsymbol{\mathrm{A}}}$,  $\dot{\boldsymbol{\mathrm{B}}}$ and $\dot{\boldsymbol{\mathrm{X}}}$.
\begin{align}
    \sum_{k}^K {\boldsymbol{\mathrm{W}}_k {\sigma}_k^p(\dot{\boldsymbol{\mathrm{X}}})} \leq \frac{1}{2}\sum_{k}^K {\boldsymbol{\mathrm{W}}_{1,k}^2{\sigma}_k^{2p}(\dot{\boldsymbol{\mathrm{A}}})}+\frac{1}{2}\sum_{k}^K{\boldsymbol{\mathrm{W}}_{2,k}^2 {\sigma}_k^{2p}(\dot{\boldsymbol{\mathrm{B}}})} \label{eq2.2.4}
\end{align}
where $\boldsymbol{\mathrm{W}}$ is the multiplication results of $\boldsymbol{\mathrm{W}}_1$ and $\boldsymbol{\mathrm{W}}_2$.
\label{the2.1}
\end{Theorem}
The proof of Theorem \ref{the2.1} can be found in the supplementary material.
\subsubsection{Overall model formulation}
Considering the effectiveness of the bilinear factor matrix norm \cite{miao2020quaternion,li2023lrrnet}, the Eq. \eqref{eq2.2.5} is reformulated as follows according to the Theorem \ref{the2.1}:
\begin{equation}
    \begin{aligned}
    &\mathop{\arg\min}_{\dot{\boldsymbol{\mathrm{A}}},\dot{\boldsymbol{\mathrm{B}}},\dot{\boldsymbol{\mathrm{D}}}} \ \ {\|\dot{\boldsymbol{\mathrm{A}}} \|}_{w,S_p,p_1=n}^p+{\|\dot{\boldsymbol{\mathrm{B}}} \|}_{w,S_p,p_1=n}^p+\beta{\| \dot{\boldsymbol{\mathrm{D}}}\|}_1 +\gamma{\|\dot{\boldsymbol{\mathrm{E}}}\|}_F^2 \\
    &s.t. \ \ \dot{\boldsymbol{\mathrm{I}}}=\dot{\boldsymbol{\mathrm{Z}}}+\dot{\boldsymbol{\mathrm{D}}}+\dot{\boldsymbol{\mathrm{E}}}, \dot{\boldsymbol{\mathrm{Z}}}=\dot{\boldsymbol{\mathrm{A}}}\dot{\boldsymbol{\mathrm{B}}}\label{eq2.2.6}
    \end{aligned}
\end{equation}
where $\dot{\boldsymbol{\mathrm{A}}}$ $\in \mathbbm{H}^{H\times r}$ and $\dot{\boldsymbol{\mathrm{B}}}$ $\in \mathbbm{H}^{r\times W}$ have the same rank $r$ to reduce the computation complexity of single factor $\dot{\boldsymbol{\mathrm{Z}}}$.
The objective function of our unified quaternion low-contrast feature learning model is defined in the following by integrating Eqs. \eqref{eq2.2.14} and \eqref{eq2.2.6}:
\begin{equation}
    \begin{aligned}
    \mathop{\arg\min}_{\dot{\boldsymbol{\mathrm{I}}},\dot{\boldsymbol{\mathrm{A}}},\dot{\boldsymbol{\mathrm{B}}},\dot{\boldsymbol{\mathrm{D}}}} \ \ &\alpha{\|\dot{\boldsymbol{\mathrm{A}}} \|}_{w,S_p,p_1=n}^p +\alpha{\|\dot{\boldsymbol{\mathrm{B}}} \|}_{w,S_p,p_1=n}^p+\beta{\| \dot{\boldsymbol{\mathrm{D}}}\|}_1 \\
    &+\gamma{\|\dot{\boldsymbol{\mathrm{E}}}\|}_F^2
    +{\|H(\nabla \dot{\boldsymbol{\mathrm{I}}}) \|}_1 +\lambda{\|\triangle \dot{\boldsymbol{\mathrm{I}}}-\triangle\dot{\boldsymbol{\mathrm{J}}}\|}_F^2\\
    &s.t. \ \ 0\leq b(\dot{\boldsymbol{\mathrm{I}}})\leq b(\dot{\boldsymbol{\mathrm{J}}}), \dot{\boldsymbol{\mathrm{G}}}=H(\nabla \dot{\boldsymbol{\mathrm{I}}})\\
    & \ \ \ \ \dot{\boldsymbol{\mathrm{I}}}=\dot{\boldsymbol{\mathrm{Z}}}+\dot{\boldsymbol{\mathrm{D}}}+\dot{\boldsymbol{\mathrm{E}}}, \dot{\boldsymbol{\mathrm{Z}}}=\dot{\boldsymbol{\mathrm{A}}}\dot{\boldsymbol{\mathrm{B}}}\label{eq2.2.6.1}
    \end{aligned}
\end{equation}
\subsubsection{Optimization}
We adopt the linearized alternating direction method with adaptive penalty (LADMAP) methods in the quaternion domain and convert \eqref{eq2.2.6} using the quaternion ADMM framework \cite{flamant2021general} as follows:
\begin{equation}
    \begin{aligned}
    \mathcal{L}= &\alpha{\|\dot{\boldsymbol{\mathrm{J}}}\|}_{w,S_p,p_1=n}^p+\frac{{\mu}_1}{2}{\|\dot{\boldsymbol{\mathrm{A}}}-\dot{\boldsymbol{\mathrm{J}}} \|}_F^2+\left \langle \dot{\boldsymbol{\mathrm{Y}}}_{3},\  \dot{\boldsymbol{\mathrm{A}}}-\dot{\boldsymbol{\mathrm{J}}}\right \rangle \\
    &+\alpha{\|\dot{\boldsymbol{\mathrm{P}}}\|}_{w,S_p,p_1=n}^p+\frac{{\mu}_1}{2}{\|\dot{\boldsymbol{\mathrm{B}}}-\dot{\boldsymbol{\mathrm{P}}} \|}_F^2+\left \langle \dot{\boldsymbol{\mathrm{Y}}}_{4},\  \dot{\boldsymbol{\mathrm{B}}}-\dot{\boldsymbol{\mathrm{P}}}\right \rangle \\
    &+\left \langle \dot{\boldsymbol{\mathrm{Y}}}_{2},\  \dot{\boldsymbol{\mathrm{Z}}}-\dot{\boldsymbol{\mathrm{A}}}\dot{\boldsymbol{\mathrm{B}}} \right \rangle+\frac{{\mu}_1}{2}{\|\dot{\boldsymbol{\mathrm{Z}}}-\dot{\boldsymbol{\mathrm{A}}}\dot{\boldsymbol{\mathrm{B}}} \|}_F^2+\beta{\| \dot{\boldsymbol{\mathrm{D}}}\|}_1 \\
    &+\gamma{\|\dot{\boldsymbol{\mathrm{E}}}\|}_F^2+\left \langle \dot{\boldsymbol{\mathrm{Y}}}_{1},\  \dot{\boldsymbol{\mathrm{I}}}-\dot{\boldsymbol{\mathrm{Z}}}-\dot{\boldsymbol{\mathrm{D}}}-\dot{\boldsymbol{\mathrm{E}}} \right \rangle\\
    &+\frac{{\mu}_1}{2}{\|\dot{\boldsymbol{\mathrm{I}}}-\dot{\boldsymbol{\mathrm{Z}}}-\dot{\boldsymbol{\mathrm{D}}}-\dot{\boldsymbol{\mathrm{E}}} \|}_F^2+{\|\dot{\boldsymbol{\mathrm{G}}} \|}_1 +\lambda{\|\triangle \dot{\boldsymbol{\mathrm{I}}}-\triangle\dot{\boldsymbol{\mathrm{L}}}\|}_F^2\\
    &+\left \langle \dot{\boldsymbol{\mathrm{Y}}}_{5},\  \dot{\boldsymbol{\mathrm{G}}}-H(\nabla \dot{\boldsymbol{\mathrm{I}}}) \right \rangle
    +\frac{{\mu}_2}{2}{\|\dot{\boldsymbol{\mathrm{G}}}-H(\nabla \dot{\boldsymbol{\mathrm{I}}}) \|}_F^2, \\
    &s.t. \ \ 0\leq b(\dot{\boldsymbol{\mathrm{I}}})\leq b(\dot{\boldsymbol{\mathrm{L}}})
    \label{eq2.2.7}
    \end{aligned} 
\end{equation}
where $\dot{\boldsymbol{\mathrm{Y}}}_{1}$ $\in \mathbbm{H}^{H \times W}$, $\dot{\boldsymbol{\mathrm{Y}}}_{2}$ $\in \mathbbm{H}^{H \times W}$,$\dot{\boldsymbol{\mathrm{Y}}}_{5}$ $\in \mathbbm{H}^{H \times W}$, $\dot{\boldsymbol{\mathrm{Y}}}_{3}$ $\in \mathbbm{H}^{H \times r}$ and $\dot{\boldsymbol{\mathrm{Y}}}_{4}$ $\in \mathbbm{H}^{r \times W}$ are the quaternion Lagrangian multipliers. $\dot{\boldsymbol{\mathrm{J}}}$ $\in \mathbbm{H}^{H \times r}$ and $\dot{\boldsymbol{\mathrm{P}}}$ $\in \mathbbm{H}^{r \times W}$ are the adjunct variables to replace $\dot{\boldsymbol{\mathrm{A}}}$ and $\dot{\boldsymbol{\mathrm{B}}}$. In addition, ${\mu}_1$ and ${\mu}_2$ are penalty factors. 
The final solution can be gained by alternately updating variables. However, too many unknown variables are coupled in Eq. \eqref{eq2.2.7}. This evidently
increases the computational complexity of the proposed model, and even results in the loss of the solution. For this reason, we left out the decomposition process temporarily, but first found the lighting suppression process for solving $\dot{\boldsymbol{\mathrm{I}}}$ and $\dot{\boldsymbol{\mathrm{G}}}$ \cite{wang2020latent}.

\textbf{Update $\dot{\boldsymbol{\mathrm{I}}}$.} Fix $\dot{\boldsymbol{\mathrm{G}}}$ and $\dot{\boldsymbol{\mathrm{Y}}}_{5}$. The subproblem of $\dot{\boldsymbol{\mathrm{I}}}$ can be solved using quaternion fast Fourier transform $\mathcal{F}(\cdot)$ under the quaternion periodic boundary condition\cite{huang2022quaternion}.
\begin{equation}
    \dot{\boldsymbol{\mathrm{I}}}=\mathcal{F}{^{-1}}({\frac{\mathcal{F}(2\lambda\triangle^H\triangle)\cdot\mathcal{F}(\dot{\boldsymbol{\mathrm{L}}})+\mu\mathcal{F}(conj(\nabla))\mathcal{F}(\dot{\boldsymbol{\mathrm{G}}})}{ \mathcal{F}(\mu\nabla^H\nabla+2\lambda\triangle^H\triangle)+eps}}).\label{eq2.2.16}
\end{equation}
\textbf{Update $\dot{\boldsymbol{\mathrm{G}}}$.} Fix $\dot{\boldsymbol{\mathrm{I}}}$ and $\dot{\boldsymbol{\mathrm{Y}}}_{5}$, The $\dot{\boldsymbol{\mathrm{G}}}$'s subproblem can be solved using Lemma \ref{lma2.2.6}.
\begin{Lemma}
    (Quaternion soft-thresholding operator \cite{xiao2018two}) Let $\dot{\boldsymbol{\mathrm{X}}} \in \mathbbm{H}$ to find the optimal solution of the problem \\
    \begin{equation}
    \begin{aligned}
        &\dot{\boldsymbol{\mathrm{X}}}=\mathop{\arg\min}_{\dot{\boldsymbol{\mathrm{X}}}}\ \ \tau{\| \dot{\boldsymbol{\mathrm{X}}} \|}_{1}
        +\frac{1}{2}\|\dot{\boldsymbol{\mathrm{X}}}-\dot{\boldsymbol{\mathrm{Y}}} \|{_F^2} \\
        &\dot{\boldsymbol{\mathrm{X}}}(:,i)=\left \{ 
    \begin{aligned}
        &\frac{{\| \dot{\boldsymbol{\mathrm{Y}}}(:,i)\|}_1-\tau}{{\| \dot{\boldsymbol{\mathrm{Y}}}(:,i)\|}_1}\dot{\boldsymbol{\mathrm{Y}}}(:,i),& &{\| \dot{\boldsymbol{\mathrm{Y}}}(:,i)\|}_1>\tau\\
        &0,& &{\rm otherwise}
    \end{aligned}
    \right.
    \end{aligned}\label{eq2.2.12}
    \end{equation}
    \label{lma2.2.6}
\end{Lemma}

\textbf{Update $\dot{\boldsymbol{\mathrm{A}}}$.}  Fix $\dot{\boldsymbol{\mathrm{B}}}$, $\dot{\boldsymbol{\mathrm{Z}}}$, $\dot{\boldsymbol{\mathrm{J}}}$, $\dot{\boldsymbol{\mathrm{Y}}}_{2}$ and $\dot{\boldsymbol{\mathrm{Y}}}_{3}$, and let the derivatives of $\mathcal{L}$ with respect to $\dot{\boldsymbol{\mathrm{A}}}$ equal to zero. Let $\dot{\boldsymbol{\mathrm{R}}}_{1}$ be equal to $(\dot{\boldsymbol{\mathrm{Z}}}+\frac{\dot{\boldsymbol{\mathrm{Y}}}_{2}}{\mu})\dot{\boldsymbol{\mathrm{B}}}^{H}+\dot{\boldsymbol{\mathrm{J}}}-\frac{\dot{\boldsymbol{\mathrm{Y}}}_{3}}{\mu}$.  The closed-form solution is obtained in $\dot{\boldsymbol{\mathrm{A}}}$'s subproblem of Eq. \eqref{eq2.2.9}.
\begin{equation}
    \begin{split}
        \dot{\boldsymbol{\mathrm{A}}}=\dot{\boldsymbol{\mathrm{R}}}_{1}{(\dot{\boldsymbol{\mathrm{B}}}\dot{\boldsymbol{\mathrm{B}}}^{H}+\dot{\boldsymbol{\mathrm{I}}}_{d})}^{-1}.
    \end{split} \label{eq2.2.9}
\end{equation}
\textbf{Update $\dot{\boldsymbol{\mathrm{B}}}$.}  Similar to $\dot{\boldsymbol{\mathrm{A}}}$'s optimization process, let the derivatives of $\mathcal{L}$ with respect to $\dot{\boldsymbol{\mathrm{B}}}$ equal to zero. Let $\dot{\boldsymbol{\mathrm{R}}}_{2}$ be equal to $\dot{\boldsymbol{\mathrm{A}}}^{H}(\dot{\boldsymbol{\mathrm{Z}}}+\frac{\dot{\boldsymbol{\mathrm{Y}}}_{2}}{\mu})+\dot{\boldsymbol{\mathrm{P}}}-\frac{\dot{\boldsymbol{\mathrm{Y}}}_{4}}{\mu}$.  The closed-form solution is obtained in $\dot{\boldsymbol{\mathrm{B}}}$'s subproblem of Eq. \eqref{eq2.2.10}.
\begin{equation}
    \begin{split}
        \dot{\boldsymbol{\mathrm{B}}}={(\dot{\boldsymbol{\mathrm{A}}}^{H}\dot{\boldsymbol{\mathrm{A}}}+\dot{\boldsymbol{\mathrm{I}}}_{d})}^{-1}(\dot{\boldsymbol{\mathrm{R}}}_{2}).
    \end{split} \label{eq2.2.10}
\end{equation}
\textbf{Update $\dot{\boldsymbol{\mathrm{Z}}}$.} Fix $\dot{\boldsymbol{\mathrm{D}}}$, $\dot{\boldsymbol{\mathrm{E}}}$,  $\dot{\boldsymbol{\mathrm{Y}}}_{1}$ and $\dot{\boldsymbol{\mathrm{Y}}}_{2}$, and let the derivatives of $\mathcal{L}$ with respect to $\dot{\boldsymbol{\mathrm{Z}}}$ equal to zero. $\dot{\boldsymbol{\mathrm{Z}}}$ can be updated below:
\begin{equation}
    \begin{split}
        \dot{\boldsymbol{\mathrm{Z}}}=(\dot{\boldsymbol{\mathrm{I}}}-\dot{\boldsymbol{\mathrm{D}}}-\dot{\boldsymbol{\mathrm{E}}}+\frac{\dot{\boldsymbol{\mathrm{Y}}}_{1}}{\mu}+\dot{\boldsymbol{\mathrm{A}}}\dot{\boldsymbol{\mathrm{B}}}-\frac{\dot{\boldsymbol{\mathrm{Y}}}_{2}}{\mu})/2. \label{eq2.2.11}
    \end{split}
\end{equation}
\textbf{Update $\dot{\boldsymbol{\mathrm{J}}}$ and $\dot{\boldsymbol{\mathrm{P}}}$.} Fix $\dot{\boldsymbol{\mathrm{A}}}$, $\dot{\boldsymbol{\mathrm{B}}}$, $\dot{\boldsymbol{\mathrm{Y}}}_{3}$ and $\dot{\boldsymbol{\mathrm{Y}}}_{4}$, $\dot{\boldsymbol{\mathrm{J}}}$ and $\dot{\boldsymbol{\mathrm{P}}}$ are solved using Lemma \ref{lma2.2.2} and \ref{lma2.2.3}. The details are found in the supplementary material.

\textbf{Update $\dot{\boldsymbol{\mathrm{D}}}$.} Fix $\dot{\boldsymbol{\mathrm{Z}}}$, $\dot{\boldsymbol{\mathrm{E}}}$ and $\dot{\boldsymbol{\mathrm{Y}}}_{1}$, the $\dot{\boldsymbol{\mathrm{D}}}$'s subproblem can be solved using Lemma \ref{lma2.2.6}.

\textbf{Update $\dot{\boldsymbol{\mathrm{E}}}$.} Fix $\dot{\boldsymbol{\mathrm{Z}}}$, $\dot{\boldsymbol{\mathrm{D}}}$ and $\dot{\boldsymbol{\mathrm{Y}}}_{1}$, the $\dot{\boldsymbol{\mathrm{E}}}$'s subproblem can be solved in a closed form as below:
\begin{equation}
    \dot{\boldsymbol{\mathrm{E}}}=\mu(\dot{\boldsymbol{\mathrm{I}}}-\dot{\boldsymbol{\mathrm{Z}}}-\dot{\boldsymbol{\mathrm{D}}})/(2\lambda+\mu).\label{eq2.2.13}
\end{equation}
More details of optimization process are provided in Algorithm \ref{algorithm2.2.1}.
\begin{algorithm}
    \caption{Quaternion low-visibility feature learning}\label{algorithm2.2.1}
    \KwIn{ The quaternion representation of the input image $\dot{\boldsymbol{\mathrm{L}}}$, the parameters ${\mu}_1$, ${\mu}_2$, $\alpha$, and $\beta$.}
    \KwOut{Optimal lighting suppressed quaternion representation $\dot{\boldsymbol{\mathrm{I}}}$ and quaternion target layer $\dot{\boldsymbol{\mathrm{D}}}$}.
    Initialize $\dot{\boldsymbol{\mathrm{G}}}{^0}$, $\dot{\boldsymbol{\mathrm{J}}}{^0}$, $\dot{\boldsymbol{\mathrm{P}}}{^0}$, ${\dot{\boldsymbol{\mathrm{Z}}}{^0}}$, $\dot{\boldsymbol{\mathrm{A}}}{^0}$, $\dot{\boldsymbol{\mathrm{B}}}{^0}$, $\dot{\boldsymbol{\mathrm{D}}}{^0}$, $\dot{\boldsymbol{\mathrm{E}}}{^0}$, $\dot{\boldsymbol{\mathrm{Y}}}{_{1}^0}$, $\dot{\boldsymbol{\mathrm{Y}}}{_{2}^0}$, $\dot{\boldsymbol{\mathrm{Y}}}{_{3}^0}$, 
 $\dot{\boldsymbol{\mathrm{Y}}}{_{4}^0}$ and $\dot{\boldsymbol{\mathrm{Y}}}{_{5}^0}$;
    
    \While{\textnormal{not converged}}{
    Fix other parameters and compute $\dot{\boldsymbol{\mathrm{I}}}{^t}$ using Eq. \eqref{eq2.2.16};

    Fix other parameters and compute $\dot{\boldsymbol{\mathrm{G}}}{^t}$ using Lemma \ref{lma2.2.6};
    
    Update $\dot{\boldsymbol{\mathrm{Y}}}{_{5}^{t+1}}=\dot{\boldsymbol{\mathrm{Y}}}{_{5}^{t}}+{\mu}_2(\dot{\boldsymbol{\mathrm{G}}}{^{t}}-H(\nabla\dot{\boldsymbol{\mathrm{I}}}{^{t}}))$;

    ${\mu}_2=\min{\{{10}^6,{\mu}_2*5\}}$;
    }
    
    Perform a normalization step to bring the illumination of $\dot{\boldsymbol{\mathrm{I}}}$ to a meaningful range.
    
    \While{\textnormal{not converged}}{
    Fix other parameters and compute $\dot{\boldsymbol{\mathrm{A}}}{^t}$ using Eq. \eqref{eq2.2.9};
     
    Fix other parameters and compute $\dot{\boldsymbol{\mathrm{B}}}{^t}$ using Eq. \eqref{eq2.2.10};

    Fix other parameters and compute $\dot{\boldsymbol{\mathrm{Z}}}{^t}$ using Eq. \eqref{eq2.2.11};

    Fix other parameters and compute $\dot{\boldsymbol{\mathrm{J}}}{^t}$ using lemma \ref{lma2.2.2} and \ref{lma2.2.3};

    Fix other parameters and compute $\dot{\boldsymbol{\mathrm{P}}}{^t}$ using lemma \ref{lma2.2.2} and \ref{lma2.2.3};

    Fix other parameters and solve $\dot{\boldsymbol{\mathrm{D}}}{^t}$ using lemma \ref{lma2.2.6};

    Fix other parameters and solve $\dot{\boldsymbol{\mathrm{E}}}{^t}$ using Eq. \eqref{eq2.2.13};
     
     Update $\dot{\boldsymbol{\mathrm{Y}}}{_{1}^{t+1}}=\dot{\boldsymbol{\mathrm{Y}}}{_{1}^{t}}+{\mu}_1(\dot{\boldsymbol{\mathrm{I}}}^t-\dot{\boldsymbol{\mathrm{Z}}}^t-\dot{\boldsymbol{\mathrm{D}}}^t-\dot{\boldsymbol{\mathrm{E}}}^t)$;

      Update $\dot{\boldsymbol{\mathrm{Y}}}{_{2}^{t+1}}=\dot{\boldsymbol{\mathrm{Y}}}{_{2}^{t}}+{\mu}_1(\dot{\boldsymbol{\mathrm{Z}}}{^{t}}-\dot{\boldsymbol{\mathrm{A}}}{^{t}}\dot{\boldsymbol{\mathrm{B}}}{^{t}})$;

       Update $\dot{\boldsymbol{\mathrm{Y}}}{_{3}^{t+1}}=\dot{\boldsymbol{\mathrm{Y}}}{_{3}^{t}}+{\mu}_1(\dot{\boldsymbol{\mathrm{A}}}{^{t}}-\dot{\boldsymbol{\mathrm{J}}}{^{t}})$;

       Update $\dot{\boldsymbol{\mathrm{Y}}}{_{4}^{t+1}}=\dot{\boldsymbol{\mathrm{Y}}}{_{4}^{t}}+{\mu}_1(\dot{\boldsymbol{\mathrm{B}}}{^{t}}-\dot{\boldsymbol{\mathrm{P}}}{^{t}})$;

       ${\mu}_1=\min{\{{10}^6,{\mu}_1*1.1\}}$;
    }
\end{algorithm}
\subsection{Quaternion adaptive unsharp masking}
The proposed quaternion adaptive unsharp masking (QAUM) method is to further enhance the image details and target saliency information and reconstruct a high-quality visible quaternion representation. 

Given the infrared $\boldsymbol{\mathrm{L}}_f$ $\in \mathbbm{R}^{H \times W}$ and the visible $\boldsymbol{\mathrm{L}}_v$ images $\in \mathbbm{R}^{H \times W \times 3}$,  the quaternion representations of these two inputs can be denoted as $\dot{\boldsymbol{\mathrm{L}}}_f$ and $\dot{\boldsymbol{\mathrm{L}}}_v$. After the quaternion low-visibility feature learning processes for the above quaternion representations, we have the lighting suppressed quaternion representations $\dot{\boldsymbol{\mathrm{I}}}_f$ and $\dot{\boldsymbol{\mathrm{I}}}_v$ and quaternion salient features $\dot{\boldsymbol{\mathrm{D}}}_f$ and $\dot{\boldsymbol{\mathrm{D}}}_v$.
The entire enhancement process in the quaternion domain for $\dot{\boldsymbol{\mathrm{I}}}_v$ can be mathematically
described as
\begin{equation}
    \dot{\boldsymbol{\mathrm{I}}}_{ve}=\dot{\boldsymbol{\mathrm{I}}}_v+\boldsymbol{\mathrm{\Lambda}}_1\odot\dot{\boldsymbol{\mathrm{D}}}_f+\boldsymbol{\mathrm{\Lambda}}_2\odot\dot{\boldsymbol{\mathrm{D}}}_v \label{eq.3.1}
\end{equation}
where $\dot{\boldsymbol{\mathrm{I}}}_{ve}$ is the enhanced visible quaternion representation, $\boldsymbol{\mathrm{\Lambda}}_1$ and $\boldsymbol{\mathrm{\Lambda}}_2$ are adaptive scaling
factor matrices unlike many existing unsharp masking algorithms that employ a fixed scaling factor for all pixels. Eq. \eqref{eq.3.1} is used in our approach to scale $\dot{\boldsymbol{\mathrm{I}}}_v$ through a pixel-wise multiplication.

We consider the scaling factor at each pixel to be computed based on the pixel intensity and saliency information of the corresponding lighting-suppressed quaternion representation. The larger the intensity or saliency value, the larger the scaling factor to be imposed.
Therefore, we construct detail-enhanced quaternion representation $\dot{\boldsymbol{\mathrm{I}}}_v$ $\in \mathbbm{H}^{H \times W}$ by extracting quaternion salient and illumination feature layers $\dot{\boldsymbol{\mathrm{D}}}_1$ and $\dot{\boldsymbol{\mathrm{D}}}_2$ from corresponding lighting suppressed quaternion representations. Here, we use a component-wise summation strategy to completely preserve the detail information of infrared-visible quaternion representations.
\begin{equation}
    \dot{\boldsymbol{\mathrm{I}}}_{ve}=\dot{\boldsymbol{\mathrm{I}}}_v+\dot{\boldsymbol{\mathrm{D}}}_1+\dot{\boldsymbol{\mathrm{D}}}_2
\end{equation}
\subsection{Quaternion Hierarchical Bayesian fusion}
A quaternion hierarchical Bayesian fusion (QHBF) model for infrared and visible image fusion is a probabilistic framework designed to integrate complementary information from IR and VIS images into a single fused image, leveraging Bayesian inference to model uncertainties, prior knowledge, and dependencies across different levels of abstraction. The objective function is formulated as:
\begin{equation}
    \begin{aligned}
    \mathop{\arg\min}_{\dot{\boldsymbol{\mathrm{F}}}} \ \ &{\|\dot{\boldsymbol{\mathrm{F}}}-\dot{\boldsymbol{\mathrm{I}}}_v \|}_{1}+\phi{\| \dot{\boldsymbol{\mathrm{F}}}-\dot{\boldsymbol{\mathrm{I}}}_f\|}_1 +\frac{1}{2}w_1{\|\nabla \dot{\boldsymbol{\mathrm{F}}}-\nabla\dot{\boldsymbol{\mathrm{I}}}_v\|}_F^2\\
    &+\frac{1}{2}w_2{\|\nabla \dot{\boldsymbol{\mathrm{F}}}-\nabla\dot{\boldsymbol{\mathrm{I}}}_f\|}_F^2,
    \label{eq2.2.22}
    \end{aligned}
\end{equation}
where $\phi$, $w_1$ and $w_2$ are real-valued parameters.
We simply substitute the two fidelity terms by 
letting $\dot{\boldsymbol{\mathrm{S}}}=\dot{\boldsymbol{\mathrm{F}}}-\dot{\boldsymbol{\mathrm{I}}}_v$, $\dot{\boldsymbol{\mathrm{T}}}=\dot{\boldsymbol{\mathrm{I}}}_f-\dot{\boldsymbol{\mathrm{I}}}_v$ where $\dot{\boldsymbol{\mathrm{T}}}$ is known. Thus, we have
\begin{equation}
    \begin{aligned}
    \mathop{\arg\min}_{\dot{\boldsymbol{\mathrm{S}}}} \ \ {\|\dot{\boldsymbol{\mathrm{S}}} \|}_{1}+\phi{\| \dot{\boldsymbol{\mathrm{T}}}-\dot{\boldsymbol{\mathrm{S}}}\|}_1 ,
    \label{eq2.2.23}
    \end{aligned}
\end{equation}
We formulate a regression problem: $\dot{\boldsymbol{\mathrm{T}}}=\dot{\boldsymbol{\mathrm{S}}}+\dot{\boldsymbol{\mathrm{Q}}}$ where $\dot{\boldsymbol{\mathrm{S}}}$ and corresponding noise $\dot{\boldsymbol{\mathrm{Q}}}$ are governed by Laplacian distribution. We can obtain the condition distribution of $\dot{\boldsymbol{\mathrm{Q}}}$ denoted as $L_q$ and the distribution of $\dot{\boldsymbol{\mathrm{S}}}$ denoted as $L_s$:
\begin{equation}
    \begin{aligned}  
        &L_q=\prod_{i,j}\frac{1}{2{\epsilon}_q}exp(\frac{\vert{\dot{q}_{i,j}}\vert}{{\epsilon}_q})\\
         &L_s=\prod_{i,j}\frac{1}{2{\epsilon}_s} exp(\frac{\vert{{\dot{s}_{i,j}}}\vert}{{\epsilon}_s}),
    \end{aligned}
\end{equation}
where ${\epsilon}_s$ and ${\epsilon}_q$ are scale parameters.
Inspired by \cite{zhao2023ddfm}, we can transform the distributions of $L_q$ and $L_s$ in a hierarchical Bayesian manner:
\begin{equation}
    \left \{
    \begin{aligned}
        &\vert\dot{q}_{i,j}\vert\mid n_{i,j} \sim \mathcal{N}(\vert\dot{q}_{i,j}\vert\mid 0,n_{i,j})\\
        &n_{i,j} \sim \mathcal{EXP}(n_{i,j},{\epsilon}_q)\\
        &\vert\dot{s}_{i,j}\vert\mid m_{i,j}\sim \mathcal{N}(\vert\dot{s}_{i,j}\vert\mid0, m_{i,j})\\
        &m_{i,j} \sim \mathcal{EXP}(m_{i,j},{\epsilon}_s)
    \end{aligned}
    \right.
\end{equation}
where $\mathcal{N}(\cdot)$ and $\mathcal{EXP}(\cdot)$ are Gaussian and exponential distributions respectively.

With the gradient regularization terms of Eq. \eqref{eq2.2.22}, the objective function is totally substituted with $\dot{\boldsymbol{\mathrm{S}}}$ and $\dot{\boldsymbol{\mathrm{T}}}$ and rewritten as 
\begin{equation}
    \mathop{\arg\min}_{\dot{\boldsymbol{\mathrm{S}}}} \ \ {\|\dot{\boldsymbol{\mathrm{S}}} \|}_{1}+\phi{\| \dot{\boldsymbol{\mathrm{Q}}}\|}_1+\frac{1}{2}w_1{\|\nabla \dot{\boldsymbol{\mathrm{S}}}\|}_F^2+\frac{1}{2}w_2{\|\nabla \dot{\boldsymbol{\mathrm{Q}}}\|}_F^2
    \label{eq2.2.24}
\end{equation}
The log likelihood of Eq. \eqref{eq2.2.24} is defined as
\begin{equation}
    \begin{aligned}
    \mathcal{L}(\dot{\boldsymbol{\mathrm{S}}})=& -\sum_{i,j}[\frac{{\vert\dot{s}_{i,j}\vert}^2}{2m_{i,j}}+\frac{{\vert\dot{q}_{i,j}
    \vert}^2}{2n_{i,j}}]\\
    &-\frac{1}{2}w_1{\|\nabla \dot{\boldsymbol{\mathrm{S}}}\|}_F^2-\frac{1}{2}w_2{\|\nabla \dot{\boldsymbol{\mathrm{Q}}}\|}_F^2
    \end{aligned} \label{eq2.2.25}
\end{equation}
In this way, we transform the objective function with quaternion variables into a maximum likelihood problem of a probability model without parameter $\phi$.

 To solve the problem of Eq. \eqref{eq2.2.25}, we adopt the $Expectation-Maximization$ (EM) algorithm \cite{dempster1977maximum} to obtain the optimal $\dot{\boldsymbol{\mathrm{S}}}$. In E-step, we formulate an expectation of log likelihood function, called the $Q$-function using the variables estimated in the $t^{th}$ iteration.
 \begin{equation}
Q(\vert\dot{\boldsymbol{\mathrm{S}}}\vert \mid{\vert\dot{\boldsymbol{\mathrm{S}}}\vert}^{(t)})=\mathbbm{E}_{M,N\vert{\dot{\boldsymbol{\mathrm{S}}}}^{(t)},{\dot{\boldsymbol{\mathrm{Q}}}}}[\mathcal{L}(\dot{\boldsymbol{\mathrm{S}}})]
 \end{equation}
In M-step, we find $\dot{\boldsymbol{\mathrm{S}}}$ corresponding to the max $Q$-function. We will show the optimization details of each step.

\textbf{E-step}. We have the expectation of the variables $\frac{1}{m_{i,j}}$ and $\frac{1}{n_{i,j}}$:
\begin{equation}
    \begin{aligned}
        &\mathbbm{E}_{m_{i,j}\mid{\vert\dot{s}_{i,j}\vert}^{(t)}}[\frac{1}{m_{i,j}}]=\sqrt{\frac{2{\vert{\dot{s}_{i,j}}^{(t)}\vert}^2}{{\epsilon}_s}}\\
        &\mathbbm{E}_{n_{i,j}\mid{\vert\dot{q}_{i,j}\vert}^{(t)}}[\frac{1}{n_{i,j}}]=\sqrt{\frac{2{\vert{\dot{q}_{i,j}}^{(t)}\vert}^2}{{\epsilon}_q}}
    \end{aligned}\label{eq2.2.26}
\end{equation}
To prove it, we set ${\widetilde{m}_{i,j}}=\frac{1}{m_{i,j}}$ and ${\widetilde{n}_{i,j}}=\frac{1}{n_{i,j}}$. Bayes' theorem gives
\begin{equation}
    \begin{aligned}
        p(\widetilde{m}_{i,j}\mid \vert\dot{s}_{i,j}\vert)&\propto p(\vert\dot{s}_{i,j}\vert,\widetilde{m}_{i,j})p(\widetilde{m}_{i,j})\\
        &\propto {\widetilde{m}_{i,j}}^{-3/2}exp[-\frac{1}{2}({\vert\dot{s}_{i,j}\vert}^2\widetilde{m}_{i,j}+\frac{2}{\widetilde{m}_{i,j}{\epsilon}_s})]
    \end{aligned}
\end{equation}
Therefore, we have
\begin{equation*}
    p(\widetilde{m}_{i,j}\mid \vert\dot{s}_{i,j}\vert)=\mathcal{IN}(\widetilde{m}_{i,j}\mid \sqrt{\frac{2{\vert\dot{s}_{i,j}\vert}^2}{{\epsilon}_s}},\frac{2}{{\epsilon}_s})
\end{equation*}
where $\mathcal{IN}(\cdot,\cdot)$ is the inverse Gaussian distribution. In a similar way, we have the posterior density of $\widetilde{n}_{i,j}$
\begin{equation*}
    p(\widetilde{n}_{i,j}\mid \vert\dot{q}_{i,j}\vert)=\mathcal{IN}(\widetilde{n}_{i,j}\mid \sqrt{\frac{2{\vert\dot{q}_{i,j}\vert}^2}{{\epsilon}_q}},\frac{2}{{\epsilon}_q})
\end{equation*}
Finally, the conditional expectation of $\frac{1}{m_{i,j}}$ and $\frac{1}{m_{i,j}}$ are the mean parameters of the corresponding inverse Gaussian distribution respectively.

\textbf{M-step}. The negative $Q$-function is minimized and the corresponding objective function is formulated as
\begin{equation}
    \mathop{\arg\min}_{\dot{\boldsymbol{\mathrm{S}}},\dot{\boldsymbol{\mathrm{Q}}}} \ \ {\Vert \boldsymbol{\mathrm{M}}\odot\dot{\boldsymbol{\mathrm{S}}}\Vert}_F^2+{\Vert \boldsymbol{\mathrm{N}}\odot\dot{\boldsymbol{\mathrm{Q}}}\Vert}_F^2+\frac{1}{2}w_1{\|\nabla \dot{\boldsymbol{\mathrm{S}}}\|}_F^2+\frac{1}{2}w_2{\|\nabla \dot{\boldsymbol{\mathrm{Q}}}\|}_F^2,
\end{equation}
where $\boldsymbol{\mathrm{M}}$ and $\boldsymbol{\mathrm{N}}$ are real-valued matrices with each element being $\mathbbm{E}_{m_{i,j}\mid{\vert\dot{s}_{i,j}\vert}^{(t)}}[\frac{1}{m_{i,j}}]$ and $\mathbbm{E}_{n_{i,j}\mid{\vert\dot{q}_{i,j}\vert}^{(t)}}[\frac{1}{n_{i,j}}]$. $\odot$ denotes element-wise multiplication of two matrices. The QADMM framework \cite{flamant2021general} is employed to solve this problem. And the hyperparameters ${\epsilon}_s$ and ${\epsilon}_q$ can be updated automatically for next iteration \cite{fu2022adaptive}.

Finally, the final fused result $\dot{\boldsymbol{\mathrm{F}}}$ is reconstructed by
\begin{equation}
    \dot{\boldsymbol{\mathrm{F}}}=\dot{\boldsymbol{\mathrm{S}}}+\dot{\boldsymbol{\mathrm{I}}}_v.
\end{equation}
\section{Experiments} \label{experiments}
To evaluate the fusion performance of our framework,
we conduct experiments on four public datasets. Section
A presents experimental settings like competing methods,
datasets, and evaluation metrics. Section B discusses the
convergence and optimal parameters of the QLCFL model. To
demonstrate the superiority of our method, Section C
presents experimental results compared with other state-of-the-art approaches. Section D presents ablation studies on the M3FD dataset \cite{liu2022target} to verify the effectiveness of our QIVIF framework.
\subsection{Experimental settings}
In our experimental settings, the stopping criterion of the QLVFL model is set to $1e-5$. Additionally, all the experiments are performed on a 2.90GHz Intel Core CPU and 16GB memory using Matlab 2016 version software.

\textbf{Datasets}. We present five types of low-visibility scenarios, including color-shift, glow effect, haze, low-light, and overexposure, as shown in Fig. \ref{fig:data}. These datasets are composed using publicly available IVIF data.
In \cite{zhang2023visible}, thirteen pairs of infrared and color visible images, each with a size of 
$328\times254$, are used in the "color-shift' dataset for IVIF tasks. In \cite{liu2022target}, twenty-two pairs of infrared and color visible images, each with a size of 
$1024\times768,$ are added to the 'glow effect' dataset. Additionally, fifteen pairs of infrared and color visible images, each with a size of 
$1024\times768$, are incorporated into the 'haze' dataset. In \cite{jia2021llvip}, five pairs of infrared and color visible images, each with a size of 
$1280\times1024$, are included in the 'low-light' dataset. Finally, in \cite{xu2020u2fusion}, ten pairs of infrared and color visible images, each with a size of 
$500\times329$, are added to the 'overexposure' dataset.

\textbf{Metrics}.
As referenced in \cite{zhang2023visible}, six objective metrics are used to quantitatively evaluate the performance of fusion quality. These include the normalized mutual information
$MI$, the entropy $EN$, the contrast-based metric $SD$, the gradient-based metrics $AG$ and $SF$, and the $Q_{abf}$. $MI$ measures the amount of shared information between the fused image and the source images (infrared and visible), evaluating how much information is preserved from both modalities. $EN$ quantifies the randomness or uncertainty (information content) in the fused image; higher 
$EN$ values generally indicate that the fused image contains more texture, complexity, and detail. $SD$ assesses the contrast or variation in pixel intensity values of the fused image. 
$SF$ measures the amount of spatial information (texture and fine details) in the image, with higher $SF$ suggesting that the fused image has more spatial details. 
$AG$ evaluates the overall gradient or edge strength in the fused image. Finally, 
$Q_{abf}$ is used to assess the overall quality of the fused image with respect to both spatial and contextual factors, considering the adaptability of the fusion method and how well it balances spatial and spectral details.

\textbf{Comparison methods}. Seven representative infrared-visible image fusion algorithms are selected as the comparison methods, which include two traditional IVIF methods LatLRR \cite{wang2020latent} and AVSHB \cite{fu2022adaptive}, general image fusion-based learning method SwinFusion \cite{ma2022swinfusion}, low-rank representation guided method LRRNet \cite{li2023lrrnet}, recent task-specific based methods MEtaFusion \cite{zhao2023metafusion} and SHIP \cite{zheng2024probing}, and a degradation-aware Network DAFusion \cite{wang2025degradation}.

\subsection{Convergence analysis and parameter sensitivity}
\begin{figure}[hbtp]
	\centering
	\includegraphics[width=1\columnwidth]{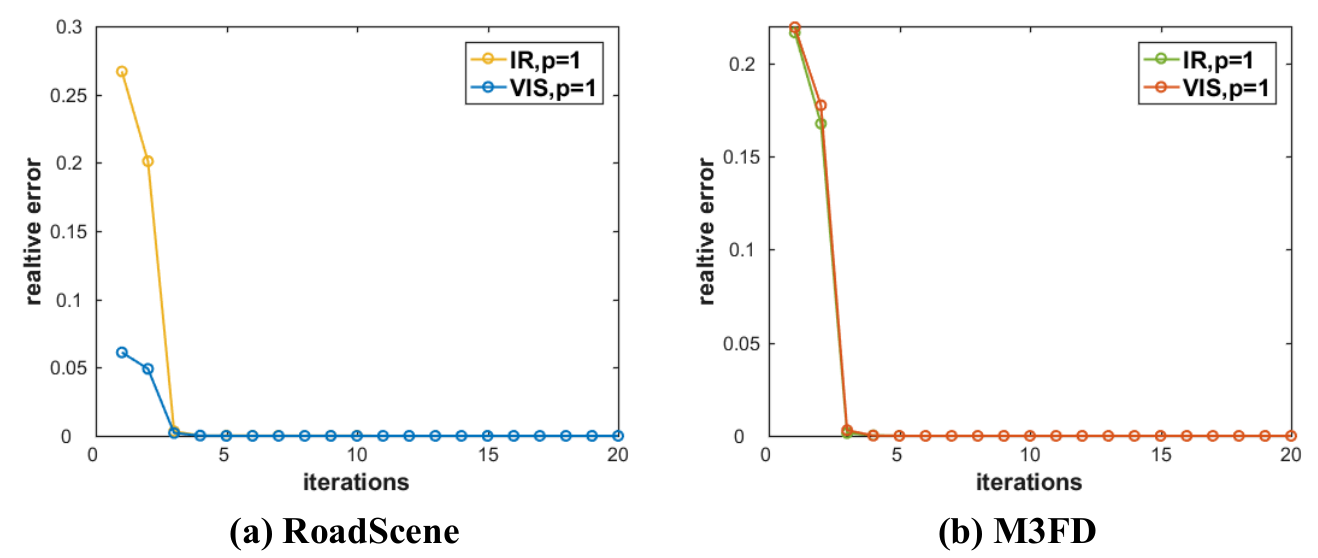}\\
	\caption{The convergence of the QLVFL model on the RoadScene \cite{xu2020u2fusion} and M3FD \cite{liu2022target} datasets.  }\label{parameter-comparison1}
\end{figure}

The convergence analysis of the QLVFL model is provided
in this subsection. Fig. \ref{parameter-comparison1} displays the evolution curves of the relative difference versus iterations of the QLVFL model on the
samples of RoadScene \cite{xu2020u2fusion} and M3FD dataset \cite{liu2022target}. The relative difference is calculated using maximum value selection. As displayed
in Fig. \ref{parameter-comparison1}, QLVFL is convergent and its total number of iterations is 20. This can sufficiently guarantee convergence.

\begin{figure}[hbtp]
	\centering
	\includegraphics[width=1\columnwidth]{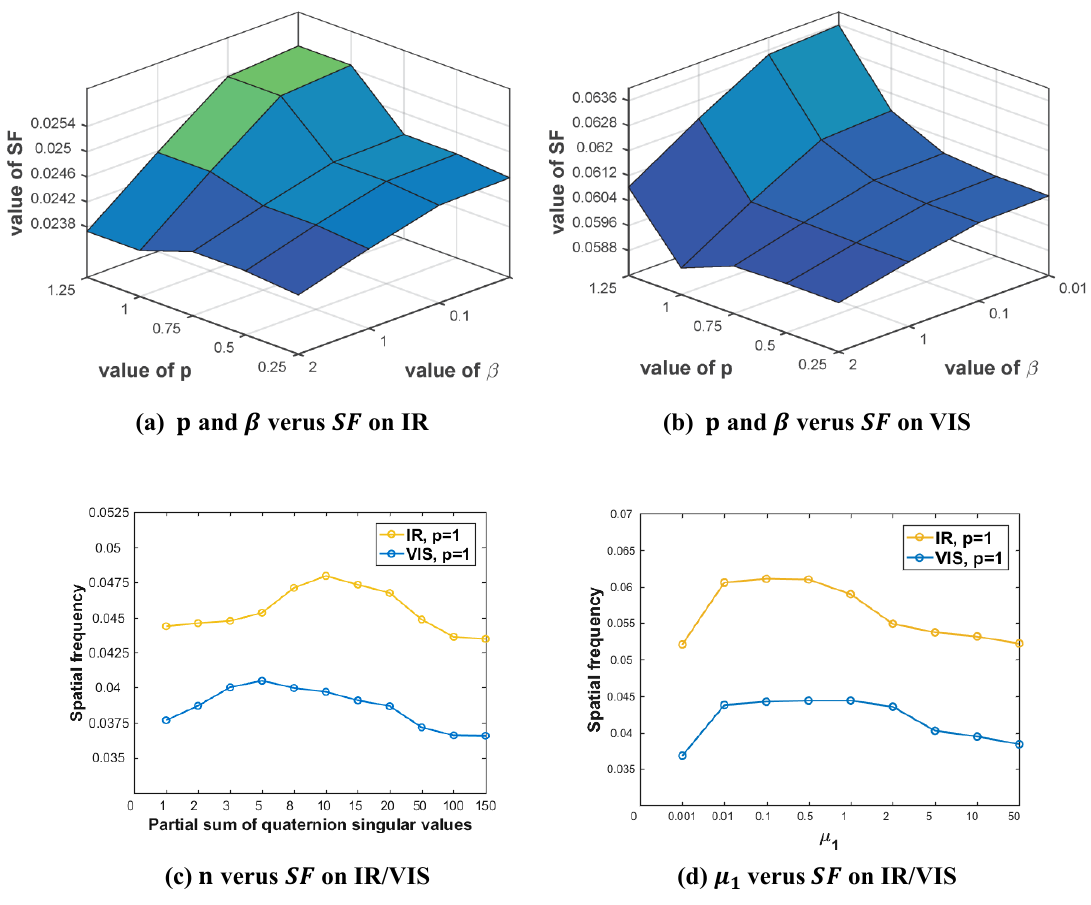}\\
	\caption{The quaternion low-visibility feature learning performance with different parameters on the M3FD dataset \cite{liu2022target}. (a) The SF value when parameter ${p}_1$ vary for the infrared and the visible on M3FD dataset \cite{liu2022target}. (b) The SF value with respect to parameter ${\mu}_1$ on M3FD dataset \cite{liu2022target}.  }\label{parameter-comparison2}
\end{figure}

In the QLVFL model, parameters $p$ and $n$ control the structure features
while parameters $\beta$ and ${\mu_1}$ handles the detail features. To quantitatively analyze the sensitivity of QLVFL versus these
parameters, Fig. \ref{parameter-comparison2} displays the SF results of QLVFL with
the samples of infrared and visible images. In Fig. \ref{parameter-comparison2} (a) and (b), $p$ and $\beta$ are set to 1 or 1.25 and 0.1 or 0.01 for the infrared or the visible. In Fig. \ref{parameter-comparison2} (c) and (d), $p_1$ is set to 10 for the infrared and ${\mu}_1$ is set to 0.5 and 0.1 for the infrared and the visible respectively.
\subsection{Infrared visible image fusion}
 \textbf{Quantitative evaluation}. The average performance of various algorithms across five challenging datasets is presented in Table \ref{table2.2.1}. A higher value indicates better performance. To effectively evaluate the objective quality of the fused images, six metrics are used for quantitative analysis. From Table \ref{table2.2.1}, it is evident that our method outperforms others in the $SD$ metric due to its strong enhancement capabilities. It suggests that our QIVIF excels in enhancing contrast and detail in the fused image.
 As shown in Table \ref{table2.2.1}, our method also surpasses the state-of-the-art in $AG$ and $SF$ and achieves the best fusion performance. $AG$ and $SF$ are both measures of spatial information. It demonstrated that our QIVIF is effective in retaining fine textures and sharp details. In terms of the $Q_{abf}$ and $MI$ indexes, our QIVIF is slightly weaker than AVSHB, as it preserves more information from the infrared images but loses some color information from the visible images.
 
 \begin{table}[hbtp]
 \small
  \caption{Quantitative comparison of proposed QIVIF and competing methods (best results are in bold).}
  \centering \resizebox{1\columnwidth}{!}{
		\begin{tabular}{|c|l|c|c|c|c|c|c|}
             \hline
	    Datasets & Methods  & $SD \uparrow$ &$SF \uparrow$ & $AG \uparrow$ & $MI \uparrow$ & $EN \uparrow$ & $Q_{abf} \uparrow$\\
			\hline
            \multirow{8}{*}{\rotatebox{90}{\textbf{Glow effect}}}&\textbf{LatLRR} \cite{wang2020latent}  & 70.12 & 0.0247 &0.0106 & 10.27 & 5.14 &0.2180\\
            
            &\textbf{AHSVB} \cite{fu2022adaptive}  & 120.12 & 0.0647 &0.0206 & 12.31 & 7.15 &\textbf{0.5180}\\
            &\textbf{SwinFusion} \cite{ma2022swinfusion}  & 133.55 & 0.0386 &0.0115 & 13.70 & 6.85 &0.4417\\
            
            &\textbf{LRRNet} \cite{li2023lrrnet}  & 98.91 & 0.0345 &0.0106 & 13.45& 6.72 &0.4613\\
           
            &\textbf{MetaFusion} \cite{zhao2023metafusion}  & 138.95 & 0.04625 &0.0177 & 14.45& 6.94 &0.4213\\
            
            &\textbf{SHIP} \cite{zheng2024probing}  & 114.13 & 0.0581 &0.0189 & \textbf{14.01} & 7.01 & 0.3996\\
            
            &\textbf{DAFusion} \cite{wang2025degradation}  & 136.67 & 0.0511 &0.0159 & 14.19 & 7.09\ & 0.5295 \\
            
            &\textbf{Ours} & \textbf{156.02} & \textbf{0.0927} &\textbf{0.0242} &14.80 & \textbf{7.40} & 0.4538\\
  \hline 
   \multirow{8}{*}{\rotatebox{90}{\textbf{Color-shift}}}&\textbf{LatLRR} \cite{wang2020latent}  & 101.56 & 0.0247 &0.0113 & 11.27 & 9.0368 &0.3180\\
   &\textbf{AHSVB} \cite{fu2022adaptive}  & 121.56 & 0.0347 &0.0221 & 12.27 & 9.14 &\textbf{0.4680}\\
            
            &\textbf{SwinFusion} \cite{ma2022swinfusion}  & 123.55 & 0.0486 &0.0315 & 10.70 & 4.85 &0.4731\\
            
            &\textbf{LRRNet} \cite{li2023lrrnet}  & 58.91 & 0.0115 &0.0101 & 11.45& 5.72 &0.4113\\
           
            &\textbf{MetaFusion} \cite{zhao2023metafusion}  & 62.91 & 0.0145 &0.0151 & 19.45& 3.72 &0.4213\\
            
            &\textbf{SHIP} \cite{zheng2024probing}  & 134.22 & 0.0481 &0.0139 & \textbf{13.01} & 8.01 & 0.2996\\
            
            &\textbf{DAFusion} \cite{wang2025degradation}  & 136.67 & 0.0491 &0.0189 & 12.19 & 6.09\ & 0.4295 \\
            
            &\textbf{Ours} & \textbf{145.02} & \textbf{0.0697} &\textbf{0.0235} &10.80 & \textbf{7.32} & 0.3238\\
            \hline
            \multirow{8}{*}{\rotatebox{90}{\textbf{Haze}}}&\textbf{LatLRR} \cite{wang2020latent}  & 71.56 & 0.0147 &0.0106 & 11.28 & 5.42 &0.3290\\
            &\textbf{AHSVB} \cite{fu2022adaptive}  & 82.56 & 0.0447 &0.0126 & 13.29 & 7.42 &\textbf{0.5290}\\
            
            &\textbf{SwinFusion} \cite{ma2022swinfusion}  & 68.55 & 0.0336 &0.025 & 14.70 & 7.85 &0.5041\\
            
            &\textbf{LRRNet} \cite{li2023lrrnet}  & 58.91 & 0.0245 &0.01006 & 12.45& 6.12 &0.4213\\
           
            &\textbf{MetaFusion} \cite{zhao2023metafusion}  & 128.91 & 0.0445 &0.0176 & 13.55& 7.72 &0.3613\\
            
            &\textbf{SHIP} \cite{zheng2024probing}  & 94.13 & 0.0581 &0.0189 & \textbf{14.01} & 7.01 & 0.3996\\
            
            &\textbf{DAFusion} \cite{wang2025degradation}  & 106.67 & 0.0511 &0.021 & 12.19 & 7.44\ & 0.5195 \\
            
            &\textbf{Ours} & \textbf{131.02} & \textbf{0.0754} &\textbf{0.0259} &13.80 & \textbf{8.40} & 0.4138\\
            \hline
            \multirow{8}{*}{\rotatebox{90}{\textbf{Low-light}}}&\textbf{LatLRR} \cite{wang2020latent}  & 102.56 & 0.0347 &0.0206 & 11.24 & 4.13 &0.1431\\
            &\textbf{AHSVB} \cite{fu2022adaptive}  & 122.56 & 0.0447 &0.0306 & 12.24 & 7.13 &\textbf{0.6431}\\
            
            &\textbf{SwinFusion} \cite{ma2022swinfusion}  & 143.55 & 0.0436 &0.0145 & 11.70 & 5.85 &0.5747\\

            &\textbf{LRRNet} \cite{li2023lrrnet}  & 93.91 & 0.0445 &0.0146 & 13.45& 6.45 &0.4644\\
           
            &\textbf{MetaFusion} \cite{zhao2023metafusion}  & 98.45 & 0.0354 &0.0154 & 13.54& 6.45 &0.4556\\
            
            &\textbf{SHIP} \cite{zheng2024probing}  & 132.13 & 0.0238 &0.0189 & \textbf{14.23} & 7.53 & 0.3223\\
            
            &\textbf{DAFusion} \cite{wang2025degradation}  & 123.67 & 0.03411 &0.024 & 14.29 & 7.78\ & 0.5235 \\
            
            &\textbf{Ours} & \textbf{184.02} & \textbf{0.0827} &\textbf{0.0224} &12.80 & \textbf{7.88} & 0.4958\\
            \hline
            \multirow{8}{*}{\rotatebox{90}{\textbf{Overexposure}}}&\textbf{LatLRR} \cite{fu2022adaptive}  & 52.56 & 0.0347 &0.0106 & 7.16 & 4.12 &0.5080\\
            &\textbf{AHSVB} \cite{fu2022adaptive}  & 122.56 & 0.0647 &0.0206 & 13.27 & 7.14 &\textbf{0.5440}\\
            
            &\textbf{SwinFusion} \cite{ma2022swinfusion}  & 133.55 & 0.0386 &0.0115 & 13.70 & 6.85 &0.5042\\

            &\textbf{LRRNet} \cite{li2023lrrnet}  & 98.91 & 0.0345 &0.0106 & 13.45& 6.72 &0.4613\\
           
            &\textbf{MetaFusion} \cite{zhao2023metafusion}  & 98.91 & 0.0345 &0.0106 & 13.45& 6.72 &0.4613\\
            
            &\textbf{SHIP} \cite{zheng2024probing}  & 114.13 & 0.0581 &0.0189 & \textbf{14.39} & 7.01 & 0.3996\\
            
            &\textbf{DAFusion} \cite{wang2025degradation}  & 136.67 & 0.0511 &0.0159 & 12.01 & 7.09\ & 0.5295 \\
            
            &\textbf{Ours} & \textbf{150.02} & \textbf{0.0727} &\textbf{0.0225} &12.75 & \textbf{7.40} & 0.4538\\
            \hline
             \multirow{8}{*}{\rotatebox{90}{\textbf{Normal-light}}}&\textbf{LatLRR} \cite{wang2020latent}  & 73.49 & 0.0647 &0.0206 & 11.23 & 6.19 &0.3360\\
             &\textbf{AHSVB} \cite{fu2022adaptive}  & 142.18 & 0.0147 &0.0206 & 12.14 & 9.134 &\textbf{0.558}\\
            
            &\textbf{SwinFusion} \cite{ma2022swinfusion}  & 123.35 & 0.0216 &0.0225 & 11.70 & 7.86 &0.4747\\
            
            &\textbf{LRRNet} \cite{li2023lrrnet}  & 108.42 & 0.0485 &0.0196 & 13.21& 5.17 &0.4325\\
           
            &\textbf{MetaFusion} \cite{zhao2023metafusion}  & 90.91 & 0.0125 &0.0126 & 11.33& 6.48 &0.4921\\
            
            &\textbf{SHIP} \cite{zheng2024probing}  & 121.22 & 0.0282 &0.0193 & \textbf{15.23} & 6.89 & 0.4997\\
            
            &\textbf{DAFusion} \cite{wang2025degradation}  & 132.89 & 0.0271 &0.0289 & 10.11 & 6.14\ & 0.3215 \\
            
            &\textbf{Ours} & \textbf{174.16} & \textbf{0.0827} &\textbf{0.0281} &14.80 & \textbf{7.40} & 0.4762\\
            \hline
  \end{tabular}}
  \label{table2.2.1}
 \end{table}
\begin{figure*}[hbtp]
	\centering
	\includegraphics[width=1\linewidth]{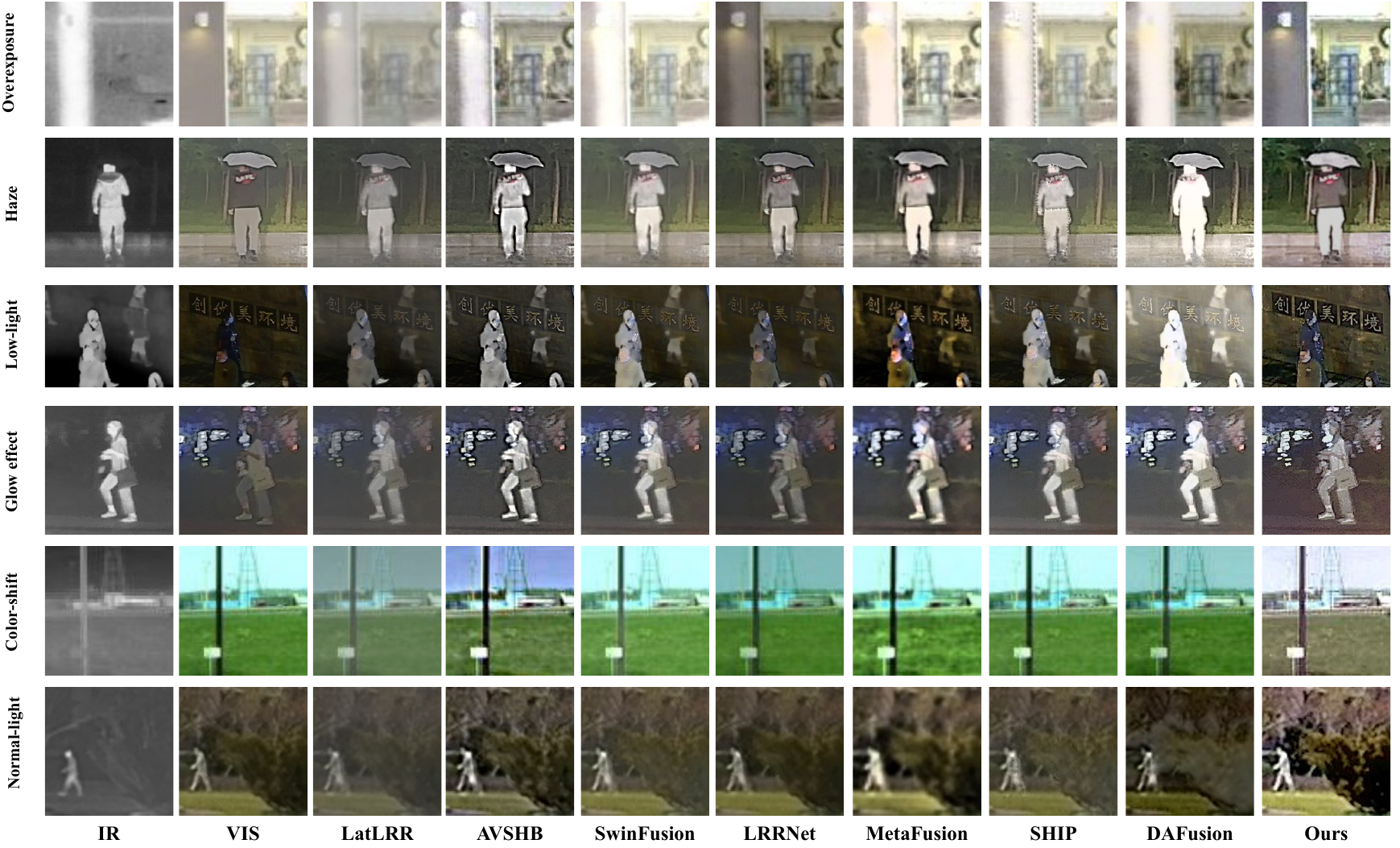}\\
	\caption{Visual comparison of infrared visible image fusion results under various low-contrast conditions.  }\label{quality-comaprison1}
\end{figure*}

\textbf{Visual evaluation}. To verify the effectiveness of our QIVIF framework under normal light and low-visibility conditions, we compare our method with seven other state-of-the-art algorithms, and discuss the fused results. Visual results are also presented to highlight areas with quality degradation. These areas are enlarged to clearly distinguish the different salient parts in the input images and their corresponding fused results.

Fig. \ref{quality-comaprison1} demonstrates the different fused results of the seven methods under various degraded conditions. They are shown in the first five rows. From the enlarged regions around all the fused results, it is evident that the results of LRRNet and SwinFusion contain many blurred pixels due to severe low-contrast and glow effects. Their feature extraction modules fail to effectively detect the salient objects and instead interfere with the visible brightness regions. Similarly, the result of MetaFusion is blurred with overexposure in the first row, as these methods do not adequately account for the fusion of low-contrast regions, leading to suboptimal detail preservation.

The last row shows the variation in fused images across the seven methods under normal light conditions. From the inputs, it is clear that the fused regions are small and adjacent to blurred regions, making it difficult to recover the details and correctly fuse the detailed parts. It is apparent that other methods fail to preserve detail information in areas such as trees and grass. In contrast, our method generates fused regions with complete detail information, ensuring high-quality fusion.

 \begin{table}[hbtp]
 \small
  \caption{Ablation study of our QIVIF framework (best results are in bold).}
   \centering \resizebox{1\columnwidth}{!}{
		\begin{tabular}{|c|c|c|c|c|c|}
        \hline
        \multicolumn{3}{|c|}{Module}& \multicolumn{3}{c|}{Metric}\\
        \hline
        QLCFL & QAUM  & QHBF &$SD \uparrow$ & $SF \uparrow$ & $EN \uparrow$\\
        \hline
        \XSolidBrush&\XSolidBrush&\Checkmark &91.7479&0.0382&6.4328\\
        \hline
        \Checkmark&\Checkmark&\XSolidBrush &116.9553&0.0691&6.9091\\
        \hline
        \Checkmark&\XSolidBrush&\Checkmark &142.5856&0.0902&7.2394\\
        \hline
        \Checkmark&\Checkmark&\Checkmark &\textbf{156.4869}&\textbf{0.1018}&\textbf{7.2797}\\
        \hline
        \end{tabular}
        }
  \label{table2.2.2}
 \end{table}
\subsection{Ablation study}
\textbf{Effectiveness of QIVIF.} To verify the advantages of the QIVIF framework, we conduct an ablation experiment on the individual modules of the proposed framework. To demonstrate the effectiveness of the QIVIF framework, we use contrast-based metric 
$SD$, gradient-based metric $SF$, and information-based metric $EN$ for performance evaluation. As clearly illustrated in Table \ref{table2.2.2}, the different settings of terms influence the fusion results to varying degrees, and the QIVIF framework yields the best quantitative results. Therefore, this experiment further validates the effectiveness of a ll the modules of QIVIF.
\begin{figure*}[hbtp]
	\centering
	\includegraphics[width=1\linewidth]{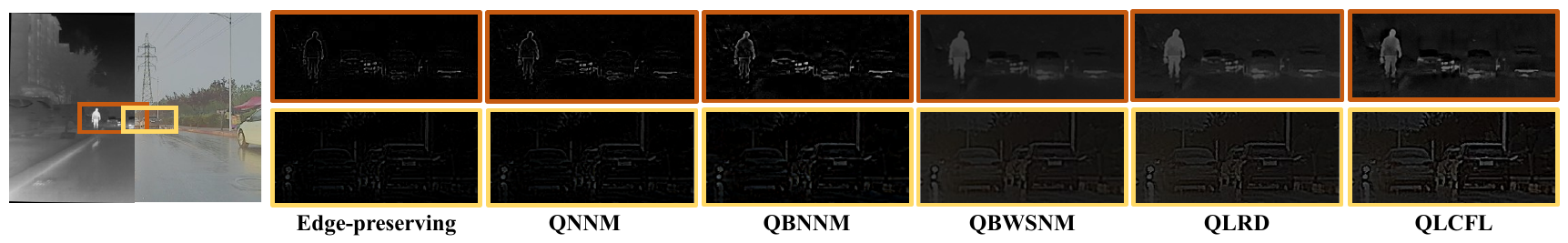}\\
	\caption{The comparison of different decomposition-based methods to validate the effectiveness of detail and salient information extraction. The first row displays the various results of $\dot{\boldsymbol{\mathrm{D}}}$ for the infrared image while the second row shows different $\dot{\boldsymbol{\mathrm{D}}}$ for the visible image.}\label{decomposition-comparison}
\end{figure*}
\begin{figure}[hbtp]
	\centering
	\includegraphics[width=1\columnwidth]{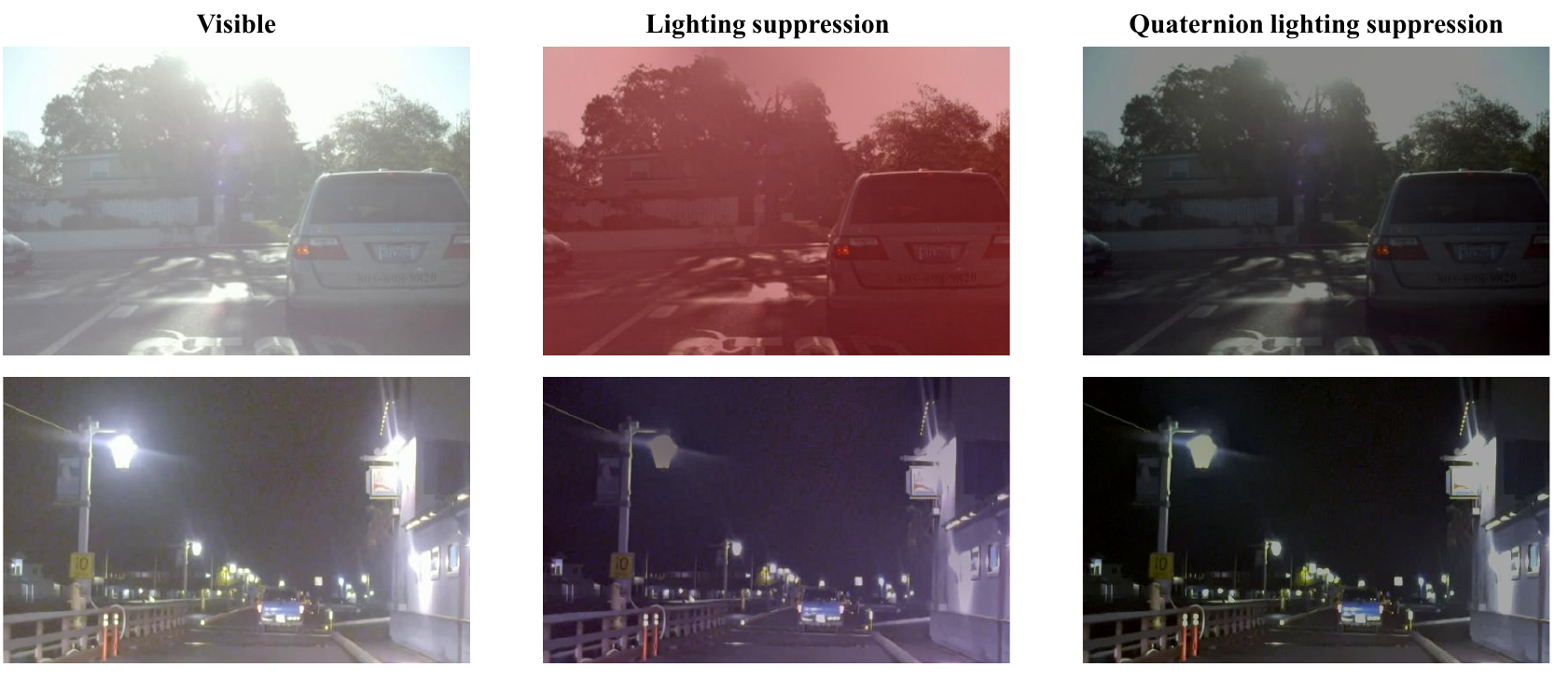}\\
	\caption{Visual comparison of lighting suppression in real and quaternion domains.  }\label{enhancement-comparison-LS}
\end{figure}

\textbf{Effectiveness of QLVFL.} To verify the advantages of the QLVFL model, we perform the ablation experiment on infrared and degraded visible images and compare various decomposition-based methods that aim to validate the effectiveness of detail and salient information extraction. The analysis of QLVFL with different model-based methods is provided, which aims to discuss how they influence the overall feature learning performance. The first row presents the results for the infrared image, while the second row displays the results for the visible image. The models include edge-preserving-based minimization, quaternion nulcear norm minimization (QNNM) , quaternion bilinear-factor nuclear norm minimization (QBNNM) \cite{miao2020quaternion}, quaternion bilinear-factor weighted Schatten-$p$ norm minimization (QBWSNM) \cite{miao2020quaternion}, QLRD and QLVFL.

We can observe how each method contributes to overall fusion performance and better understand the strengths of the proposed framework in extracting and integrating relevant details as clearly illustrated in Fig. \ref{decomposition-comparison}.  Furthermore, this experiment verifies the effectiveness of quaternion lighting suppression without color casts in Fig. \ref{enhancement-comparison-LS}.

\section{Conclusion} \label{conclusion}
We proposed a quaternion-based infrared-visible image fusion framework that is entirely in the quaternion domain to comprehensively extract complementary features from both infrared and visible modalities. This enabled robust enhancement and fusion under various low-visibility conditions. The framework introduced a quaternion low-visibility feature learning module, which iteratively learns salient thermal targets and fine-grained texture details from both infrared and visible images in challenging conditions. To enhance the details of a degraded visible image, the framework incorporated a quaternion adaptive unsharp masking method, which injected salient and detailed information to reconstruct the visible image with balanced illumination. Additionally, a quaternion hierarchical Bayesian fusion model was proposed to integrate infrared saliency and enhanced visible details, ensuring high-visibility outputs that preserve thermal targets and maintain natural color balance. Extensive experiments across diverse datasets demonstrated that our framework outperforms state-of-the-art IVIF methods in both qualitative and quantitative metrics, particularly under challenging low-visibility conditions

\bibliographystyle{IEEEtran}
\bibliography{base}

\vfill

\end{document}